\documentclass[conference]{IEEEtran}
\IEEEoverridecommandlockouts
\usepackage{cite}
\usepackage{amsmath,amssymb,amsfonts}
\usepackage{algorithmic}
\usepackage{graphicx}
\usepackage{textcomp}

\usepackage{algorithm}
\usepackage{xcolor}
\usepackage{multirow}
\usepackage{caption}
\usepackage{subcaption}  
\usepackage{threeparttable} 
\usepackage{framed}
\usepackage{pgfplots}
\pgfplotsset{compat=1.18}
\usepackage{colortbl}
\usepackage{tcolorbox} 
\usepackage{float}

\usepackage{enumitem}
\usepackage{booktabs}
\usepgfplotslibrary{groupplots}
\usepackage{cuted} 
\tcbuselibrary{breakable}
\tcbuselibrary{skins} 

\setlist[itemize]{beginpenalty=0, endpenalty=0} 

\usepackage{pgfkeys}
    \newenvironment{customlegend}[1][]{%
        \begingroup
        \csname pgfplots@init@cleared@structures\endcsname
        \pgfplotsset{#1}%
    }{%
        \csname pgfplots@createlegend\endcsname
        \endgroup
    }%
    \def\addlegendimage{\csname pgfplots@addlegendimage\endcsname}
\definecolor{chartpurple}{RGB}{178,178,250}
\definecolor{chartpink}{RGB}{254,178,178}

\definecolor{cGreen}{RGB}{0,150,0}
\definecolor{brown}{RGB}{139,64,0}

\definecolor{rebuttalGreen}{RGB}{97,153,101}

\definecolor{rebuttalOrange}{RGB}{199,125,72}

\definecolor{rebuttalBlue}{RGB}{55,154,211}

\newcommand{\ournameAbbr}{{BGC }}

\def\BibTeX{{\rm B\kern-.05em{\sc i\kern-.025em b}\kern-.08em
    T\kern-.1667em\lower.7ex\hbox{E}\kern-.125emX}}
\begin{document}
\title{Backdoor Graph Condensation
}
\author{
    \IEEEauthorblockN{Jiahao Wu$^{1,2}$, Ning Lu$^{1,3}$, Zeyu Dai$^{1,2}$, Kun Wang$^4$, Wenqi Fan$^{2}$, Shengcai Liu$^{1}$*\thanks{* Corresponding author. Email: {liusc3@sustech.edu.cn}.}, Qing Li$^{2}$, Ke Tang$^{1}$}
    \IEEEauthorblockA{$^1$ Guangdong Provincial Key Laboratory of Brain-Inspired Intelligent Computation, SUSTech, Shenzhen, China}
    \IEEEauthorblockA{$^2$ The Hong Kong Polytechnic University, Hong Kong, China}
     \IEEEauthorblockA{$^3$ Hong Kong University of Science and Technology, Hong Kong, China}
     \IEEEauthorblockA{$^4$ University of Science and Technology of China, Hefei, China}
}

\maketitle

\begin{abstract}
Graph condensation has recently emerged as a prevalent technique to improve the training efficiency for graph neural networks (GNNs). It condenses a large graph into a small one such that a GNN trained on this small synthetic graph can achieve comparable performance to a GNN trained on the large graph. However, while existing graph condensation studies mainly focus on the best trade-off between graph size and the GNNs' performance (model utility), they overlook the security issues of graph condensation. To bridge this gap, we first explore backdoor attack against the GNNs trained on the condensed graphs.

We introduce an effective backdoor attack against graph condensation, termed BGC. This attack aims to (1) preserve the condensed graph quality despite trigger injection, and (2) ensure trigger efficacy through the condensation process, achieving a high attack success rate. Specifically, BGC consistently updates triggers during condensation and targets representative nodes for poisoning.
Extensive experiments demonstrate the effectiveness of our attack. BGC achieves a high attack success rate (close to 1.0) and good model utility in all cases. Furthermore, the results against multiple defense methods demonstrate BGC's resilience under their defenses. Finally, we analyze the key hyperparameters that influence the attack performance. {Our code is available at: https://github.com/JiahaoWuGit/BGC.} 

\end{abstract}


\section{Introduction}
Graph neural networks (GNNs) have been widely deployed in various fields involving graph-structured data, such as social computing~\cite{tkde2023ssslSocialComputing,graphPro24}, drug discovery~\cite{drug-discovery2021,jiatong23LLMDrug} and recommendation~\cite{jiahao2023DConRec,wu-et-al:DcRec_cikm22,wenqi23LLMRec}. Large-scale graphs are instrumental in achieving state-of-the-art GNNs performance~\cite{gcond2024surveyEmory,junfengFan2023NeurIPS,wenqi2024GraphLLM,dyj2024wwwCompanion}. However, the vast scale of these graphs imposes considerable demands on storage and computational resources.

\begin{figure*}[]
\centering
\vskip -0.08in
{\includegraphics[width=0.88\linewidth]{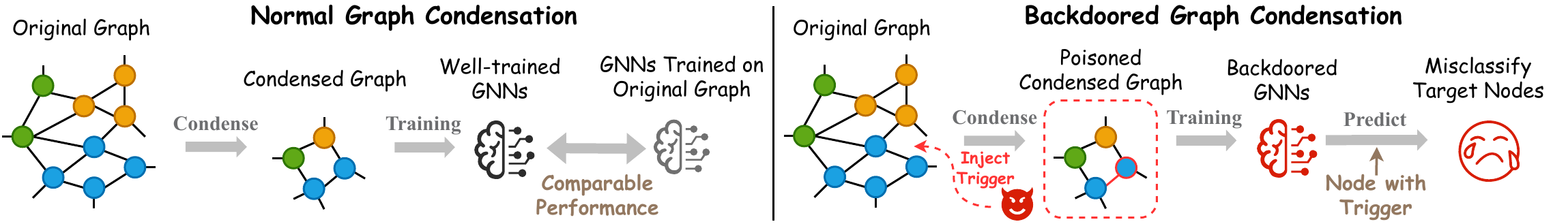}}
\vskip -0.08in
\caption{Normal Graph Condensation \textit{vs} Backdoored Graph Condensation. }
\vskip -0.25in
\label{fig:intro-gc-backdoor}
\end{figure*}

Graph condensation~\cite{corr2023GCondEigen,doscond-kdd2022,fang2024exgcWebConference,sfgc2023nips,kdd2023LiteGNTK,kdd2024GCSR,wangling2023fastGC} is an emerging solution to the aforementioned challenges. Graph condensation compresses a large graph into a smaller one, enabling GNNs trained on it to achieve performance comparable to those trained on the original graph, as shown in Figure~\ref{fig:intro-gc-backdoor}. For instance, GCond~\cite{iclr2022gcond} condenses the Reddit dataset (153,932 training nodes) into only 154 synthetic nodes. GCond reduces the number of training nodes by 99.09\% while retaining 95.3\% of the original test performance. 
Regarding the remarkable achievements, we expect graph condensation to be provided as a service~\cite{gcond2024surveyUQ,gcond2024surveyZJU,ndss2023doorping}. Such service can ease storage and computational demands for researchers and organizations.

{Despite the success, the parameter optimization process in condensation inherits neural networks' security  vulnerabilities~\cite{sp2017NeuralCleanse,iclr2015adverse,icml2015scalableAttack,shijie2023untargetedAttack,wsdm2023JunfengFang,liangbo24wwwCompanion}, while existing graph condensation research}~\cite{iclr2022gcond,sfgc2023nips,kdd2023LiteGNTK,wangling2023fastGC,doscond-kdd2022} primarily focuses on achieving best trade-off between the graph size reduction and the model utility. The security issues of graph condensation haven't been investigated. {This is highly concerning given that (i) graph-structured data is increasingly used in security-sensitive domains such as malware analysis~\cite{ijcai2019p522}, fraud detection~\cite{ndss2019fraudDetec}, and drug discovery~\cite{drugDiscovery2018}, and (ii) data providers like Scale AI facilitate machine learning model development, introducing potential security risks while their trustworthiness is in question.} To bridge this research gap, we propose the task of backdoor graph condensation, aiming to investigate the security issue of graph condensation.

\begin{figure}
    \centering
    \setlength\tabcolsep{2pt}
    \begin{tabular}{c}
        \resizebox{1\linewidth}{!}{
            \begin{tikzpicture}
    \begin{customlegend}[
        legend columns=3,
        legend style={
            anchor=north,             
            draw=none,               
            fill=white,               
            text depth=0pt,           
            font=\footnotesize,       
            /tikz/every even column/.append style={column sep=0.3cm},
            legend image code/.code={
                \draw[fill=##1, draw=white, /tikz/.cd,yshift=-0.3em]
                (0cm,0cm) rectangle (0.35em,0.5em); 
            }
        },
        legend entries={
            \textsc{Clean Model},
            \textsc{Naive Poison},
            \textsc{Our Method},
        }
        ] 
        \addlegendimage{ybar,ybar legend, 
        fill={rgb, 255:red,197;green,96;blue,83},
        draw=black}
        \addlegendimage{ybar,ybar legend, 
        fill={rgb, 255:red,97;green,153;blue,101},
        draw=black}
        \addlegendimage{ybar,ybar legend, 
        fill={rgb, 255:red,102;green,138;blue,217},
        draw=black}

        \end{customlegend}
\end{tikzpicture}
        }\\
    \begin{tabular}{cc}
    \begin{subfigure}{0.49\linewidth}
        \resizebox{1.0\linewidth}{!}{
            \scalebox{1.0}{
\begin{tikzpicture}
\begin{axis}[
    ybar,
    enlargelimits=0.9, 
    enlarge y limits={upper,value=0.05},  
    ylabel={CTA (\%)},
    ylabel style={font=\huge},  
    xlabel={Cora},
    xlabel style={font=\huge},  
    symbolic x coords={CM, NP, BGC},
    xtick=\empty,
    xticklabel style={font=\huge, rotate=0},
    yticklabel style={font=\huge},  
    grid=none,  
    ymajorgrids=true,  
    axis background/.style={
        fill={rgb, 255:red,228;green,228;blue,237},
        fill opacity=0.7  
    },
    grid style={color=white},  
    axis line style={color=white},  
    tick style={color=white},  
    bar width=28pt,
    ymin=50, ymax=85,  
    xtick pos=bottom,  
    every axis plot/.append style={
        fill,
        draw=none
    },
]
    \addplot+[
        fill={rgb, 255:red,197;green,96;blue,83},
        draw=black,
        error bars/.cd,
        y dir=both,
        y explicit,
        error bar style={draw=black, line width=1pt},  
        error mark 
    ] coordinates {
        (CM,81.23) 
    };
    \addplot+[
        fill={rgb, 255:red,97;green,153;blue,101},
        draw=black,
        error bars/.cd,
        y dir=both,
        y explicit,
        error bar style={draw=black, line width=1pt},  
        error mark options={rotate=90, mark size=2pt, draw=black}  
    ] coordinates {
        (NP,64.30) 
    };
    \addplot+[
        fill={rgb, 255:red,102;green,138;blue,217},
        draw=black,
        error bars/.cd,
        y dir=both,
        y explicit,
    ] coordinates {
        (BGC,80) 
    };
\end{axis}
\end{tikzpicture}
}

        }
    \end{subfigure}
    &
    \begin{subfigure}{0.483\linewidth}
        \resizebox{1.0\linewidth}{!}{
            \scalebox{1.0}{
\begin{tikzpicture}
\begin{axis}[
    ybar,
    enlargelimits=0.9, 
    enlarge y limits={upper,value=0.05},  
    ylabel={CTA (\%)},
    ylabel style={font=\huge},  
    xlabel={Citeseer},
    xlabel style={font=\huge},  
    symbolic x coords={CM, NP, BGC},
    xtick=\empty,
    xticklabel style={font=\huge, rotate=0},
    yticklabel style={font=\huge},  
    grid=none,  
    ymajorgrids=true,  
    axis background/.style={
        fill={rgb, 255:red,228;green,228;blue,237},
        fill opacity=0.7  
    },
    grid style={color=white},  
    axis line style={color=white},  
    tick style={color=white},  
    bar width=28pt,
    ymin=0, ymax=75,  
    xtick pos=bottom,  
    every axis plot/.append style={
        fill,
        draw=none
    },
]
    \addplot+[
        fill={rgb, 255:red,197;green,96;blue,83},
        draw=black,
        error bars/.cd,
        y dir=both,
        y explicit,
        error bar style={draw=black, line width=1pt},  
        error mark 
    ] coordinates {
        (CM,71.57) 
    };
    \addplot+[
        fill={rgb, 255:red,97;green,153;blue,101},
        draw=black,
        error bars/.cd,
        y dir=both,
        y explicit,
        error bar style={draw=black, line width=1pt},  
        error mark options={rotate=90, mark size=2pt, draw=black}  
    ] coordinates {
        (NP,34.36) 
    };
    \addplot+[
        fill={rgb, 255:red,102;green,138;blue,217},
        draw=black,
        error bars/.cd,
        y dir=both,
        y explicit,
    ] coordinates {
        (BGC,70) 
    };
\end{axis}
\end{tikzpicture}
}
        }
    \end{subfigure}
\end{tabular}
\\
    \end{tabular}
    \vskip -0.07in
    \caption{Attack Performance Comparison: Naively Poisoning Condensed Graphs \textit{vs} Our Method. Here, CTA denotes the test accuracy on clean graphs. {Naive Poison degrades condensed graph quality, thus reducing GNN utility.}}
    \label{fig:intro-naive-vs-bgc}
    \vskip -0.23in
\end{figure}

Graph backdoor attacks~\cite{adversarialGNNs_2019iclr,dai2023www-unnoticeable,wang2023brave,wsdm2023JunfengFang}, which aim to investigate the vulnerability of GNNs, are closely related to the new task. These attacks utilize triggers to poison input samples, causing GNNs to produce adversary-predetermined malicious outputs. Traditional backdoor methods inject triggers directly into the graph during GNN training. However, such approaches are ineffective for condensed graphs due to their small size, which makes triggers easy to detect and remove. For example, Reddit's condensed graph contains only 154 nodes, rendering abnormal edge detection straightforward and limiting trigger efficacy~\cite{ndss2023doorping,dai2023www-unnoticeable,gta2021usenix}. Furthermore, directly injecting triggers into condensed graphs significantly compromises the GNN utility. As illustrated in Figure~\ref{fig:intro-naive-vs-bgc},
direct injection into the condensed graph (Naive Poison) substantially undermines the GNNs utility (CTA), compared to GNNs trained on a clean graph (Clean Model). Therefore, directly adapting existing graph backdoor attacks to poison the condensed graph is ineffective, raising a critical question: 
     \textit{{How to inject backdoors into GNNs without directly poisoning the condensed graph?}}

To address this, we first summarize the primary objectives for backdoor attacks against graph condensation: 1) high GNN utility, which aims to maintain the condensed graph quality to preserve the GNN utility; and 2) high attack effectiveness, which emphasizes enduring trigger influence throughout condensation to ensure the effectiveness. However, \textbf{pursuing these two objectives is challenging.} {Specifically, the condensed graph quality heavily relies on the original graph and the condensation method, complicating trigger design. Additionally, preserving trigger effectiveness is intractable. Consistent updates to the condensed graph reduce trigger efficacy, while static triggers fail to poison dynamic nodes.}

To overcome the challenges, we propose the first \textbf{B}ackdoor attack against \textbf{G}raph \textbf{C}ondensation, \textbf{BGC}. We envision the attacker as the graph condensation service provider as shown in Figure~\ref{fig:intro-gc-backdoor}.
Specifically, instead of direct injection into the condensed graph, BGC inserts triggers into the original large graph and {optimize them iteratively throughout the condensation process to ensure their retention}. Further, to preserve condensed graph quality, trigger attachments are limited to a controlled number of high-influence nodes~\cite{liu2023dreamDD,xu2023distillDDMassive}. We validate BGC’s effectiveness through extensive experiments on four benchmark datasets and four representative condensation methods. Empirical results show that our method BGC achieves high model utility and achieves a promising attack success rate. Besides, ablation studies confirm that BGC is robust in different settings. The evaluation result with two representative graph defense methods indicates that they are invalid in defending against BGC. Our contributions are:
\begin{itemize}[leftmargin=*]
    \item \textbf{Objective.} We introduce backdoor graph condensation, a task that embeds malicious information into condensed graphs to backdoor GNNs, uncovering vulnerabilities and associated risks in graph condensation.
    \item  \textbf{Methodology.} We propose the first backdoor attack against graph condensation, where we devise a module to select representative nodes for trigger injection in the original graph and update the triggers throughout the condensation process. This enables effective backdoor attacks on GNNs trained on the condensed graph.
    \item \textbf{Experiment.} We evaluate BGC across various settings, demonstrating its consistent effectiveness (near 100\% attack success rate and high GNNs utility). Besides, the evaluation with two leading graph defense methods indicates that our attacks can surpass these defense methods and module ablation study helps justify the design effectiveness.
\end{itemize}

\section{Preliminaries}
\label{sec:preliminaries}
In this section, we introduce some notations and present the preliminaries on graph condensation and graph backdoor attacks. 
We denote a graph $\mathcal{G}=\{\mathbf{A},\mathbf{X},\mathbf{Y}\}$, where $\mathbf{A}\in \mathbb{R}^{N\times N}$ is the adjacency matrix, $N$ is the number of nodes, $\mathbf{X}\in \mathbb{R}^{N\times d}$ is the $d$-dimensional node feature matrix and $\mathbf{Y}\in \{0,1,...,C-1\}^N$ denotes the node labels over $C$ classes. Here, $\mathbf{A}_{ij}=1$ if node $v_i$ and node $v_j$ are connected; otherwise $\mathbf{A}_{ij}=0$. We denote the node set by $\mathcal{V}=\{v_1,...,v_N\}$. In this paper, we focus on node classification tasks.

\subsection{Graph Condensation}
\textbf{Goal}: Graph condensation~\cite{iclr2022gcond,wangling2023fastGC,jiahao23TF-DCon,sfgc2023nips} aims to  learn a small, synthetic graph dataset $\mathcal{S}=\{\mathbf{A}',\mathbf{X}',\mathbf{Y}'\}$. where $\mathbf{A}'\in \mathbb{R}^{N'\times N'}$, $\mathbf{X}'\in \mathbb{R}^{N'\times d}$, $\mathbf{Y}'\in \{0,1,...,C-1\}^{N'}$ and $N'\ll N$. A GNN trained on the condensed graph $\mathcal{S}$ is expected to achieve comparable performance to the one trained on the original graph $\mathcal{G}$. 

\textbf{Formulation}: Graph condensation can be formulated as a bi-level optimization problem~\cite{doscond-kdd2022,iclr2022gcond,zhaoICLR2021DC,jiahao2023DConRec,wangling2023fastGC}, iteratively updating the synthetic graph $\mathcal{S}$ and the model parameters ${\theta}$. The optimization objective is formulated as:

\scalebox{0.88}{\parbox{1.1\linewidth}{
\begin{align}
    {\min_{{\mathcal{S}}}} \mathcal{L}(f({\mathcal{G}}|{\theta}_{\mathcal{S}}), {\bf Y}) \; 
    \textit{s.t.} \; {\theta}_{\mathcal{S}}=\underset{{\theta}}{\arg \min }\mathcal{L}(f(\mathcal{S}|{\theta}), {\bf Y'}),
    \label{eq:bi_level_cond}
\end{align}
}}
where ${f(\cdot|{\theta}_{\mathcal{S}})}$ denotes the GNN model parameterized with ${\theta_{\mathcal{S}}}$, ${\theta}_{\mathcal{S}}$ denotes the GNN is trained on the synthetic graph $\mathcal{S}$, and $\mathcal{L}$ is the task-related loss utilized for GNN training (i.e., cross entropy loss). In pursuit of this objective, different condensation methods may have different designs for the optimization of the synthetic graph and one of the most prevalent methods adopted by previous works is the gradient matching mechanism~\cite{iclr2022gcond,doscond-kdd2022,zhaoICLR2021DC,fang2024exgcWebConference}. Concretely, they endeavor to minimize the discrepancy between GNN gradients $w.r.t.$ the original graph $\mathcal{G}$ and $w.r.t.$ the condensed graph $\mathcal{S}$~\cite{fang2024exgcWebConference}. Therefore, the GNNs trained on $\mathcal{S}$ will converge to similar states and achieve comparable performance to those trained on $\mathcal{G}$.


\subsection{Graph Backdoor Attack}
\label{sec:graph-backdoor-preliminary}
\textbf{Goal}: The attacker injects triggers into the training data of the target GNN model and attaches triggers to target nodes at the test time, leading to misclassification on target nodes while maintaining normal behavior for clean nodes without triggers. At the setting of graph backdoor attack~\cite{dai2023www-unnoticeable}, the data is available for the attacker while the information of the target GNN models is unknown to the attacker. The effectiveness of a backdoor attack is typically evaluated by attack success rate (ASR) and clean test accuracy (CTA)~\cite{ndss2023doorping,iclr2024rethinking,dai2023www-unnoticeable}. The ASR measures its success rate in misleading GNN to predict the given triggered samples to the target label. The CTA evaluates the utility of the model given clean samples.

\textbf{Formulation}: Given a clean graph $\mathcal{G}=\{\mathbf{A},\mathbf{X},\mathbf{Y}\}$ with node set $\mathcal{V}$, the goal of graph backdoor attack~\cite{dai2023www-unnoticeable} is to learn an adaptive trigger generator $f_g:v_i\rightarrow g_i$ and effectively select a set of nodes $\mathcal{V}_P$ within budget to attach triggers and labels so that the GNN $f$ trained on the poisoned graph $\mathcal{G}_P$ will classify the test node attached with the trigger to the target class $y_t$ {by solving}:\\
\scalebox{0.9}{\parbox{1.1\linewidth}{
\begin{align}
    &\min_{\mathcal{V}_P, {\theta}_{g}}  \sum_{v_i \in {\mathcal{V}_U}} l(f(a(\mathcal{G}_C^i,g_i|{{\theta}^*})), y_t),\notag\\
    &s.t.~{\theta}^*=\mathop{\arg\min}_{{\theta}}\sum_{v_i \in \mathcal{V}_{C}}l(f(\mathcal{G}_C^i|{{\theta}}), y_i)+ \sum_{v_i \in \mathcal{V}_P}l(f(a(\mathcal{G}_C^i,g_i)|{\theta}), y_t), \notag\\
    & \forall v_i \in \mathcal{V}_P \cup \mathcal{V}_{T},~~|g_i| < \Delta_{g}, \text{ and } |\mathcal{V}_P| \leq \Delta_{P},
    \label{eq:problem-form-graph-backdoor}
\end{align}
}}\\
where we have:
\begin{itemize}[leftmargin=*,nosep,label={}]
    \item {$\mathcal{V}_{C}=\mathcal{V}\backslash \mathcal{V}_P$: the clean node set;}
    \item {$\mathcal{V}_T$: test node set;}
    \item {$\mathcal{V}_U\subseteq\mathcal{V}$: node set for updating triggers.}
    \item {${\theta}_g$: parameters of the adaptive trigger generator $f_g$;}
    \item {$g_i$: trigger of node $v_i$;}
    \item {$\Delta_g$: budget of trigger size;}
    \item {$\Delta_P$: budget of poisoned node set $\mathcal{V}_P$;}
    \item {$\mathcal{G}^i_C$: clean computation graph of node $v_i$;}
    \item {$a(\cdot)$: the operation of trigger attachment;}
    \item {$l(\cdot)$: node-level form of the cross-entropy loss;}
\end{itemize}

\noindent{While the prediction is given based on the computation graph of the node, the clean prediction on node $v_i$ can be written as $f(\mathcal{G}_C^i|\theta)$. For a node $v_i$ attached with the adaptive trigger $g_i$, the predictive lable will be given by $f_{\theta}(a(\mathcal{G}^i_C,g_i))$.}

To be consistent with Eq~\ref{eq:bi_level_cond},
we have following definition:
\begin{align*}
\mathcal{L}\left(f({\mathcal{G}}|{\theta}),\mathbf{Y}\right)=\sum_{v_i \in \mathcal{V}} l(f(\mathcal{G}_C^i|{\theta}),y_i),
\end{align*}

\section{Problem Formulation}
\label{sec:problem_form}
\textbf{Attack Settings.} We consider a practical scenario, where the attacker is envisioned as the malicious graph condensation provider, supplying condensed graphs. Besides, we assume that the attacker is accessible to the dataset but lack the information of the target model. This stems from that the provider solely delivering condensed graphs, without knowledge of the model to be trained.

\textbf{Attacker's Goal.} The attacker's goal is to inject malicious information into the condensed graph dataset and consequently backdoor the GNNs trained on the condensed graph. However, as discussed in the previous section, the size of the condensed graph is significantly smaller than the original graph,
which will result that the direct injection of triggers into the condensed graphs renders those triggers easily detectable and it can significantly influence the utility of GNN~\cite{dai2023www-unnoticeable,ndss2023doorping}. Therefore, we turn to a more practical strategy, injecting triggers into the original graph throughout the condensation process to carry out the attacks.

\textbf{Formulation of Backdoor Graph Condensation.} Given a clean graph $\mathcal{G}=\{\mathbf{A},\mathbf{X},\mathbf{Y}\}$ with node set $\mathcal{V}$, we aim to learn an adaptive trigger generator $f_g:v_i\rightarrow g_i$ and effectively select a set of nodes $\mathcal{V}_P\subset \mathcal{V}$ within budget to attach triggers and labels. Then, we can denote the poisoned graph by $\mathcal{G}_P=\{\mathbf{A}^P,\mathbf{X}^P,\mathbf{Y}^P\}$. The condensed graph produced from $\mathcal{G}_P$ is denoted by $\mathcal{S}=\{\mathbf{A}',\mathbf{X}',\mathbf{Y}'\}$. Thus, we learn the trigger generator by solving the following problem:
\begin{equation}
\begin{aligned}
    &\min_{\mathcal{G}_P} \;\hat{\mathcal{L}}(f({\mathcal{G}_P}|\theta_{\mathcal{S}^*}), {\bf Y}^P)\\
    &\textit{ s.t. } \;\theta_{\mathcal{S}^*}=\underset{\theta}{\arg \min } \; \mathcal{L}\left(f({\mathcal{S}}|\theta), {\bf Y'}\right),\\
    &\textit{ s.t. } \; 
    \mathcal{S}^* =  \arg\min_{{\mathcal{S}}}\; \mathcal{L}'(f({\mathcal{S}}|\theta_{\mathcal{S}}), {\bf Y}^P), \;
\end{aligned}
\label{eq:trilevel_level_cond}
\end{equation}
where {$\hat{\mathcal{L}}$, ${\mathcal{L}}$, and ${\mathcal{L}'}$ are the graph-level form of loss, which will be elaborated in the following paragraphs.} 

{$\hat{\mathcal{L}}$ is cross-entropy loss and it is utilized to optimize trigger generator, aiming to} mislead GNN $f$ trained on $\mathcal{S}$ to classify the test nodes attached with the trigger to the target class $y_t$ and it is formulated as following: 

\scalebox{0.95}{\parbox{1.0\linewidth}{
\begin{equation}
    \begin{aligned}
        &\hat{\mathcal{L}}(f({\mathcal{G}_P}|\theta_{\mathcal{S}^*}), {\bf Y}^P) = \sum_{v_i \in {\mathcal{V}_U}} l(f(a(\mathcal{G}_C^i,g_i)|\theta_{\mathcal{S}^*}), y_t),\\
        & \forall v_i \in \mathcal{V}_P \cup \mathcal{V}_{T}, ~~|g_i| < \Delta_{g}\text{ and } |\mathcal{V}_P| \leq \Delta_{P},
    \end{aligned}
    \label{eq:trilevel_level_cond_notes}
\end{equation}
}}
where $\theta_g$, $y_t$, $f_g$, $\mathcal{G}^i_C$, $v_i$, $a(\cdot)$, $l(\cdot)$, $\mathcal{V}_T$ and $\mathcal{V}_U$ are coherent with Eq.(\ref{eq:problem-form-graph-backdoor}). 

{$\mathcal{L}$ is cross-entropy loss and it is utilized to train GNN $f$ on the condensed graph $\mathcal{S}$}:
\begin{equation}
    \mathcal{L}\left(f({\mathcal{S}}|\theta), {\bf Y'}\right)=\sum_{v_i \in \mathcal{V}'} l(f(\mathcal{G}_C^i|\theta),y_i),
\end{equation}
where $\mathcal{V}'$ is the node set of synthetic graph $\mathcal{S}$.

{$\mathcal{L}'$ is the loss for graph condensation and we introduce the gradient matching mechanism~\cite{iclr2022gcond,doscond-kdd2022,wangling2023fastGC,sfgc2023nips} in this paper}. Therefore, the optimization of $\mathcal{S}$ can be re-written as:

\scalebox{0.83}{\parbox{1.15\linewidth}{
\begin{align}    
    \mathcal{L}'(f({\mathcal{G}^P}|\theta_{\mathcal{S}}), {\bf Y}^P)=D(\nabla_{\theta}\mathcal{L}(f({\mathcal{S}}|\theta_t),{\bf Y'}),\nabla_{\theta}\mathcal{L}(f({\mathcal{G}}|\theta_t),{\bf Y})),
\end{align}
}}
where $D(\cdot,\cdot)$ is a distance function, $T$ is the number of steps of the whole GNN's training trajectory.
\section{Methodology}
\label{sec:methodology}
\subsection{Overview}
\label{sec:methodology-overview}
\textbf{Motivation.} As shown in Figure~\ref{fig:intro-naive-vs-bgc}, directly injecting triggers into the condensed graph can significantly compromise the GNN utility (CTA). Therefore, this approach cannot carry out effective backdoor attacks. Instead, we propose to inject triggers into the original graph. 
Examining the bi-level pipeline of graph condensation~\cite{iclr2022gcond,doscond-kdd2022,wangling2023fastGC,zjiahao24unlearnGraph}, we observe that the condensed graphs are consistently optimized. {If triggers are pre-defined and remain unchanged, they may not be preserved throughout the updating process~\cite{ndss2023doorping,gta2021usenix}, failing to inject malicious information into the condensed graph. Thus, we search to devise an advanced attack, where triggers are updated during condensation, preserving their effectiveness. Further, to avoid undermining the quality of the condensed graph, we set a budget to limit the number of poisoned nodes in the original graph. To maximize the effectiveness of triggers within this limit, we propose to attach triggers to the representative nodes.}

{The interplay of consistently updated triggers and representative poisoning ensures potent, stealthy attacks ($\approx 100\%$ success rate) without degrading model performance on clean data. These two techniques together challenges the perceived security of condensed graphs, revealing risks in outsourced ML services, urging defenses tailored to dynamic graph condensation and rethinking node vulnerability.}

\begin{figure}[]
\centering
\vskip -0.17in
{\includegraphics[width=0.9\linewidth]{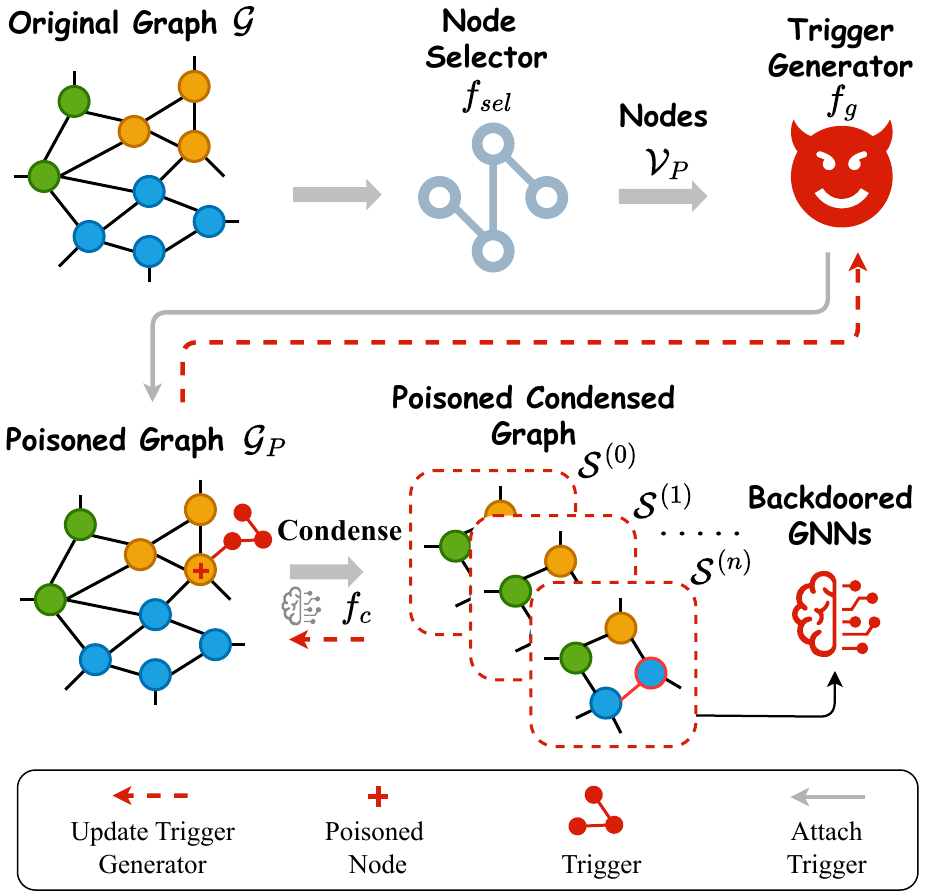}}
\vskip -0.05in
\caption{Overview of Backdoor Graph Condensation.}
\vskip -0.25in
\label{fig:frame-iter}
\end{figure}

\textbf{Overview of BGC.} In this section, we present the details of our proposed attack BGC, illustrated in Figure~\ref{fig:frame-iter}. {{Initially}}, a poisoned node selector $f_{sel}$ is trained on the original graph $\mathcal{G}$ and selects representative nodes as poisoned nodes $\mathcal{V}_P$ through a metric that measures the representativeness of each node~\cite{xu2023distillDDMassive,liu2023dreamDD}. {{Subsequently}}, an adaptive trigger generator $f_g$ generates triggers for the selected poisoned nodes $\mathcal{V}_P$, making up part of the poisoned graph $\mathcal{G}_P$. {{Finally}}, the poisoned condensed graph $\mathcal{S}_P$ is generated based on $\mathcal{G}_P$.

\textbf{Optimization.} As formulated in Eq~\ref{eq:trilevel_level_cond}, generating effective triggers and corresponding backdoored condensed graphs is a tri-level optimization problem, which is both challenging and computationally expensive. To address this issue, we update the triggers before updating the condensed graph at each epoch. This significantly reduces the number of iterations and ensures consistent updates to the toxic triggers for sustained effectiveness. Since the information of the target model is unknown, the trigger generator and condensed graph are optimized towards successfully attacking a surrogate GCN model $f_c$. The optimization is summarized in Algorithm~\ref{algorithm:BGC}.

\subsection{Poisoned Node Selection}
\label{sec:poison-node-select-methodology}
In this section, we delve into poisoned node selection. Previous studies~\cite{liu2023dreamDD,xu2023distillDDMassive} indicate that representative samples play a pivotal role in the gradient of dataset condensation, exerting a substantial influence on the quality of condensed datasets. Additionally, injecting too many triggers into the graphs can affect the GNN utility. Therefore, to facilitate successful backdoor attacks, we limit the budget of poisoning and we propose to select representative nodes to poison. Then, we introduce the selection method in detail.

Specifically, we train a GCN model $f_{sel}$ with the original graph $\mathcal{G}=\{\mathbf{A},\mathbf{X},\mathbf{Y}\}$ to obtain the representations of nodes:
\begin{equation}
    \label{eq:GCN-represent}
    \mathbf{H}_{sel}=\text{GCN}_{sel}(\mathbf{A},\mathbf{X}),\quad \hat{\mathbf{Y}}=\text{softmax}(\mathbf{W}_{sel}\cdot\mathbf{H}_{sel}),
\end{equation}
where $\mathbf{W}_{sel}$ is the weight matrix for classification. The training objective of $f_{sel}$ can be formulated as:
\begin{equation}
    \label{eq:train-gcn}
    \min_{{\theta}_{sel},\mathbf{W}_{sel}}\sum_{v_i\in \mathcal{V}}l(\hat{y_i},y_i),
\end{equation}
where ${\theta}_{sel}$ is the parameters of node selector $f_{sel}$, $l(\cdot)$ is the cross-entropy loss, $y_i$ is node label and $\hat{y}_i$ is the prediction.

To guarantee the diversity of the representative nodes~\cite{liu2023dreamDD,dai2023www-unnoticeable}, we separately apply K-Means to cluster on each class $c$. 
Nodes nearer to the cluster centroid are more representative and nodes nearest to the centroid may have a high degree. Assigning malicious labels to high-degree nodes could lead to substantial performance degradation since the negative effects will be propagated to a wide range of neighbors. Therefore, we adopt the metric that can balance the representativeness and the negative effects on GNN utility~\cite{dai2023www-unnoticeable}. The metric score can be calculated as follows:
\begin{equation}
    \label{eq:metric-node-select}
    m(v_i)=||\mathbf{h}^k_i-\mathbf{h}^k_c||_2+\lambda\cdot deg(v_i^k),
\end{equation}
where $\lambda$ is the balance hyperparameter, $\mathbf{h}^k_c$ is the representation of the centroid of the $k$-th cluster, and $\mathbf{h}^k_i$ is the representation of a node $v_i^k$, belonging to $k$-th cluster. After getting the scores, we select the nodes with top-$n$ highest scores in each cluster. $n=\frac{\Delta_P}{(C-1)K}$, where $C$ is the number of classes, $K$ is the number of clusters, {and $\Delta_P$ is the attack budget introduced in Section~\ref{sec:graph-backdoor-preliminary}}.
\subsection{Trigger Generation}
\label{sec:trigger-generation}
Given the poisoned node set $\mathcal{V}_P$, we utilize the generator $f_g$ to generate the triggers for nodes in $\mathcal{V}_P$ to poison the graph. Since different nodes possess different structural neighbors, the most effective topology of the triggers may vary. Therefore, $f_g$ takes the node features and graph structure as input to generate not only the node features but also the structure of the triggers. Therefore, the generator $f_g$ is optimized during BGC's process, not the triggers.

Specifically, we adopt a GCN model to encode the node features and graph structure into representations as following:
\begin{equation}
    \label{eq:trigger-gen}
    \mathbf{H}=\text{GCN}_g(\mathbf{A},\mathbf{X}).
\end{equation}
Then, the node features and structure of the trigger for node $v_i$ is generated as following:
\begin{equation}
    \label{eq:trigger-gen2}
\mathbf{X}^g_i=\mathbf{W}_f\cdot\mathbf{h}_i,\quad\mathbf{A}_i^g=\mathbf{W}_a\cdot\mathbf{h}_i,
\end{equation}
where $h_i$ is the representation of $v_i$. $\mathbf{W}_f$ and $\mathbf{W}_a$ are the learnable parameters of $f_g$ for feature and structure generation, respectively. $\mathbf{X}^g_i\in\mathbb{R}^{|g_i|\times d}$ is the synthetic features of the trigger nodes, where $|g_i|$ is the size of the generated trigger and $d$ is the dimension of features. $\mathbf{A}^g_i\in\mathbb{R}^{|g_i|\times|g_i|}$ is the adjacency matrix of the generated trigger. Since the graph structure is discrete, we follow previous studies~\cite{hubara2016binarized,dai2023www-unnoticeable} to binarize the continuous adjacency matrix $\mathbf{A}^g_i$ in the forward computation while using the continuous adjacency matrix value in backward propagation. With the generated trigger $g_i=(\mathbf{X}_i^g,\mathbf{A}_i^g)$, we link it to node $v_i\in\mathcal{V}_P$ and assign target class label $y_t$ to build the backdoored graph $\mathcal{G}_P$ to generate the condensed graph $\mathcal{S}$ for model training. During inference, the generator $f_g$ will generate triggers for test nodes to lead the backdoored GNN to misclassify them as target class $y_t$. 

\begin{algorithm}[]
\caption{Optimization of \ournameAbbr}
\label{algorithm:BGC}
    \begin{algorithmic}[1]
    \REQUIRE The original graph $\mathcal{G}=\{\mathbf{A},\mathbf{X},\mathbf{Y}\}$, $T$, $M$
    \ENSURE The condensed graph $\mathcal{S}$, adaptive generator trigger $f_g$
    \STATE Initialize poisoned graph $\mathcal{G}_P=\mathcal{G}$;
    \STATE Randomly initialize the condensed graph $\mathcal{S}$, $\boldsymbol{\theta}_{sel}$ for poisoned node selector $f_{sel}$ and $\boldsymbol{\theta}_g$ for adaptive trigger generator $f_g$;
    \STATE Select poisoned nodes $\mathcal{V}_P$ based on {Eq~\ref{eq:metric-node-select}} and assign class $y_t$ as labels of $\mathcal{V}_P$;
    \WHILE{update condensed graph $\mathcal{S}$}
        \STATE Initialize the surrogate model $\boldsymbol{\theta}_c$;
        \FOR{t=$1,2,\dots{},T$}
        \STATE Update the surrogate model $\boldsymbol{\theta}_c$ by based on Eq~\ref{eq:update-surrogate};
        \ENDFOR
        \FOR{t=$1,2,\dots{},M$}
        \STATE Update the trigger generator $\boldsymbol{\theta}_g$ based on Eq~\ref{eq:update-trigger-gen};
        \ENDFOR
        \STATE Attach updated triggers $\{g_i\}$ to the selected nodes in $\mathcal{V}_P$ to build the poisoned graph $\mathcal{G}_P$.
        \STATE Update the condensed graph $\mathcal{S}$ based on Eq~\ref{eq:update-condensed-graph};
    \ENDWHILE
    \RETURN $\mathcal{S}$ and $f_g$
    \end{algorithmic}
\end{algorithm}
\vskip -0.2in
\subsection{Optimization}
\label{sec:trigger-optim}
Since the target model $f$ is invisible to attackers, we propose to optimize the trigger generator $f_g$ and generate condensed graph $\mathcal{S}$ to successfully attack the surrogate GCN model $f_c$. In this section, we elaborate on the optimization of condensed graph $\mathcal{S}$, surrogate model $f_c$ and the trigger generator $f_g$. 

\textbf{Surrogate Model.} Given the condensed graph $\mathcal{S}=\{\mathbf{A}',\mathbf{X}',\mathbf{Y}'\}$, the surrogate GCN model $f_c$ is trained by:
\begin{equation}
    \label{eq:surrogate-model-train}
    \min_{{\theta}} \mathcal{L}_c = \;\mathcal{L}\left(f_c({\mathcal{S}}|\theta), {\bf Y'}\right),
\end{equation}
where ${\theta}$ is the parameters of $f_c$.

\textbf{Trigger Generator.} Given the selected poison nodes $\mathcal{V}_P$ and the surrogate GCN model ${f_c(\cdot|{\theta}_{\mathcal{S}^*})}$ optimally trained on $\mathcal{S}$, we can update the trigger generator $f_g$ to mislead the surrogate model to classify nodes with trigger to label $y_t$ by:
\begin{equation}
    \label{eq:trigger-gen-opt}
    \min_{{\theta}_g}\mathcal{L}_g= \sum_{v_i \in {\mathcal{V}_U}} l(f_c(a(\mathcal{G}_C^i,g_i)|{\theta}_{\mathcal{S}^*}), y_t),
\end{equation}
where $\mathcal{G}^i_C$ indicates the clean computation graph of node $v_i$, $g_i$ is the trigger of $v_i$, $a(\cdot)$ represents the attachment operation, $l(\cdot)$ is the cross-entropy loss, $y_t$ is the target class, ${\theta}_g$ is the parameters of trigger generator $f_g$ and nodes in $\mathcal{V}_U\subseteq\mathcal{V}$ are randomly sampled from $\mathcal{V}$ to ensure that the attacks can be effective for various types of target nodes.
\begin{table}[]
\centering
\caption{Dataset Statistics. Cora and Citeseer are transductive datasets while Flickr and Reddit are inductive datasets.}
\vskip -0.1in
\resizebox{0.9\linewidth}{!}{ 
\begin{tabular}{|cc|rrrr|}
\hline
\multicolumn{2}{|c|}{Dataset} & \multicolumn{1}{c}{Cora} & \multicolumn{1}{c}{Citeseer} & \multicolumn{1}{c}{Flickr} & \multicolumn{1}{c|}{Reddit} \\ \hline
\multicolumn{2}{|c|}{\#Nodes}       & 2,708 & 3,327 & 89,250   & 232,965   \\
\multicolumn{2}{|c|}{\#Edges}       & 5,429 & 4,732 & 1,166,243 & 57,307,946 \\
\multicolumn{2}{|c|}{\#Classes}     & 7    & 6    & 40      & 210      \\
\multicolumn{2}{|c|}{\#Features}    & 1,433 & 3,703 & 500     & 602      \\\hline
\multirow{3}{*}{Split} & Train    & 140  & 120  & 44,625   & 153,932   \\
                       & Val. & 500  & 500  & 22,312   & 23,699    \\
                       & Test     & 1,000 & 1,000 & 22,312   & 55,334    \\ \hline
\end{tabular}
}
\label{tab:data-statistic}
\vskip -0.2in
\end{table}
\begin{table*}[]
\centering
\vskip -0.05in
\caption{Model Utility (CTA) and Attack Performance (ASR). Across all settings, our method achieves high model utility, with CTA scores closely approaching C-CTA scores, and achieves a high attack success rate (ASR).}
\vskip -0.05in
\resizebox{0.9\linewidth}{!}{ 
\begin{threeparttable}
\newcolumntype{G}{>{\columncolor{gray!20}}c} 
\begin{tabular}{|c|c|cGcG|cGcG|}
\hline
\multirow{2}{*}{Datasets} &
  \multirow{2}{*}{Ratio (r)} &
  \multicolumn{4}{c|}{DC-Graph} &
  \multicolumn{4}{c|}{GCond} \\ \cline{3-10} 
 &
   &
  C-CTA &
  CTA &
  C-ASR &
  ASR &
  C-CTA &
  CTA &
  C-ASR &
  ASR \\ \hline
\multirow{3}{*}{Cora} &
  1.30\% &
  75.17 (0.65) &
  75.90 (0.94) &
  12.51 (1.03) &
  100.0 (0.00) &
  81.33 (0.62) &
  81.23 (0.24) &
  11.23 (1.31) &
  100.0 (0.00) \\
 &
  2.60\% &
  75.97 (0.61) &
  75.00 (0.51) &
  13.23 (0.61) &
  100.0 (0.00) &
  81.27 (0.33) &
  80.67 (0.52) &
  13.42 (0.06) &
  100.0 (0.00) \\
 &
  5.20\% &
  78.30 (0.43) &
  78.43 (0.31) &
  12.12 (0.13) &
  100.0 (0.00) &
  80.53 (0.73) &
  80.70 (0.50) &
  11.78 (0.85) &
  100.0 (0.00) \\ \hline
\multirow{3}{*}{Citeseer} &
  0.90\% &
  68.37 (0.90) &
  70.27 (0.50) &
  15.49 (1.34) &
  100.0 (0.00) &
  71.43 (0.33) &
  71.57 (1.32) &
  16.65 (0.19) &
  100.0 (0.00) \\
 &
  1.80\% &
  69.10 (0.62) &
  69.37 (0.94) &
  15.04 (0.40) &
  100.0 (0.00) &
  72.03 (0.17) &
  71.03 (0.09) &
  14.64 (0.58) &
  100.0 (0.00) \\
 &
  3.60\% &
  70.23 (0.05) &
  70.00 (0.22) &
  17.12 (0.73) &
  100.0 (0.00) &
  71.20 (0.70) &
  70.60 (0.51) &
  16.18 (0.49) &
  100.0 (0.00) \\ \hline
\multirow{3}{*}{Flickr} &
  0.10\% &
  45.49 (0.43) &
  46.48 (0.21) &
  2.54 (0.34) &
  99.98 (0.02) &
  46.85 (0.10) &
  46.54 (0.08) &
  2.18 (0.43) &
  99.83 (0.07) \\
 &
  0.50\% &
  46.37 (0.10) &
  46.44 (0.13) &
  2.68 (0.90) &
  99.25 (0.65) &
  46.62 (0.53) &
  47.15 (0.08) &
  2.25 (0.51) &
  99.97 (0.02) \\
 &
  1.00\% &
  47.14 (0.03) &
  46.77 (0.30) &
  2.63 (0.07) &
  99.11 (0.40) &
  46.91 (0.41) &
  46.84 (0.09) &
  2.21 (0.35) &
  99.77 (0.06) \\ \hline
\multirow{3}{*}{Reddit} &
  0.05\% &
  85.88 (0.08) &
  85.38 (0.42) &
  0.46 (0.01) &
  99.90 (0.03) &
  88.86 (0.06) &
  88.50 (0.27) &
  0.45 (0.01) &
  99.84 (0.14) \\
 &
  0.10\% &
  89.25 (0.21) &
  89.14 (0.05) &
  0.49 (0.01) &
  99.93 (0.02) &
  89.20 (0.13) &
  90.37 (0.22) &
  0.47 (0.01) &
  99.99 (0.01) \\
 &
  0.20\% &
  91.15 (0.07) &
  90.38 (0.42) &
  0.46 (0.00) &
  99.90 (0.03) &
  90.10 (0.26) &
  90.40 (0.41) &
  0.45 (0.02) &
  99.06 (0.91) \\ \hline\hline
\multirow{2}{*}{Datasets} &
  \multirow{2}{*}{Ratio (r)} &
  \multicolumn{4}{c|}{GCond-X} &
  \multicolumn{4}{c|}{GC-SNTK} \\ \cline{3-10} 
 &
   &
  C-CTA &
  CTA &
  C-ASR &
  ASR &
  C-CTA &
  CTA &
  C-ASR &
  ASR \\ \hline
\multirow{3}{*}{Cora} &
  1.30\% &
  77.67 (0.59) &
  76.30 (1.35) &
  14.03 (0.52) &
  100.0 (0.00) &
  81.24 (0.50) &
  79.73 (1.40) &
  13.73 (1.44) &
  98.30 (1.53) \\
 &
  2.60\% &
  78.70 (0.90) &
  78.10 (0.71) &
  11.93 (0.68) &
  100.0 (0.00) &
  80.50 (0.70) &
  79.10 (1.04) &
  13.05 (0.29) &
  99.77 (0.15) \\
 &
  5.20\% &
  80.40 (1.28) &
  79.40 (0.36) &
  12.02 (0.73) &
  100.0 (0.00) &
  80.12 (0.32) &
  79.67 (0.42) &
  13.17 (0.15) &
  98.13 (1.65) \\ \hline
\multirow{3}{*}{Citeseer} &
  0.90\% &
  73.76 (0.21) &
  73.03 (0.50) &
  17.64 (0.34) &
  100.0 (0.00) &
  60.68 (1.04) &
  60.20 (3.64) &
  15.16 (1.34) &
  100.0 (0.00) \\
 &
  1.80\% &
  72.07 (0.61) &
  72.40 (0.57) &
  17.24 (0.26) &
  100.0 (0.00) &
  62.33 (2.06) &
  63.03 (2.64) &
  16.92 (0.56) &
  100.0 (0.00) \\
 &
  3.60\% &
  72.16 (0.00) &
  72.13 (0.83) &
  15.94 (0.35) &
  100.0 (0.00) &
  63.44 (1.30) &
  63.37 (2.06) &
  16.41 (0.25) &
  100.0 (0.00) \\ \hline
\multirow{3}{*}{Flickr} &
  0.10\% &
  45.60 (0.80) &
  46.15 (0.47) &
  2.67 (0.06) &
  98.26 (1.19) &
  46.10 (0.10) &
  46.32 (0.09) &
  2.51 (0.26) &
  99.98 (0.00) \\
 &
  0.50\% &
  46.68 (0.20) &
  45.21 (0.44) &
  2.43 (0.35) &
  99.58 (0.35) &
  46.23 (0.10) &
  46.02 (0.21) &
  2.58 (0.32) &
  99.98 (0.00) \\
 &
  1.00\% &
  45.74 (0.31) &
  45.62 (0.17) &
  2.49 (0.09) &
  95.51 (1.69) &
  46.01 (0.20) &
  45.50 (0.06) &
  2.60 (0.09) &
  99.98 (0.00) \\ \hline
\multirow{3}{*}{Reddit} &
  0.05\% &
  87.09 (0.10) &
  87.47 (0.21) &
  0.48 (0.00) &
  99.89 (0.07) &
  OOM &
  OOM &
  OOM &
  OOM \\
 &
  0.10\% &
  88.42 (0.47) &
  89.14 (0.51) &
  0.46 (0.00) &
  99.58 (0.30) &
  OOM &
  OOM &
  OOM &
  OOM \\
 &
  0.20\% &
  89.96 (0.14) &
  90.09 (0.21) &
  0.46 (0.01) &
  97.60 (1.42) &
  OOM &
  OOM &
  OOM &
  OOM \\ \hline
\end{tabular}
\begin{tablenotes}
    \item C-CTA (\%): Clean Test Accuracy with the Clean GNNs.\qquad\quad\, CTA (\%): Clean Test Accuracy with the Backdoored GNNs.
    \item C-ASR (\%): Attack Success Rate with the Clean GNNs.\;\,\qquad\quad ASR (\%): Attack Success Rate with the Backdoored GNNs.
\end{tablenotes}
\end{threeparttable}
}
\label{tab:model-utility-attack-performance}
\vskip -0.2in
\end{table*}

\textbf{Condensed Graph.} Given the selected poison nodes $\mathcal{V}_P$ and trigger generator $f_g$ parameterized by ${\theta}_g$, we can generate the poisoned graph $\mathcal{G}_P=\{\mathbf{A}^P,\mathbf{X}^P,\mathbf{Y}^P\}$, based on which we can obtain $\mathcal{S}$ via:
\begin{equation}
    \label{eq:condense-graph}
    \min_{{\mathcal{S}}}\;\mathcal{L}_s = \mathcal{L}'({\mathcal{G}^P}|{\theta}_{\mathcal{S}}), {\bf Y}^P),
\end{equation}
where ${\theta}_{\mathcal{S}}$ is the parameters of surrogate GCN model $f_c$ trained on $\mathcal{S}$ and $\mathcal{L}'$ is the loss function for graph condensation.

Combining Eq~\ref{eq:surrogate-model-train}, Eq~\ref{eq:trigger-gen-opt} and Eq~\ref{eq:condense-graph}, the tri-level optimization problem could be re-written as:
\begin{equation}
\begin{aligned}
    &\min_{{\theta}_g} \;\mathcal{L}_g({\theta}_{\mathcal{S}^*({\theta}_g)},\mathcal{S}^*({\theta}_g),{\theta}_g)\\
    &\textit{ s.t. } \;{\theta}_{\mathcal{S}^*}=\underset{{\theta}}{\arg \min } \; \mathcal{L}_c({\theta},\mathcal{S}^*({\theta}_g),{\theta}_g),\\
    &\textit{ s.t. } \; 
    \mathcal{S}^* =  \arg\min_{{\mathcal{S}}}\; \mathcal{L}_s(\mathcal{S},{\theta}_g). \;
\end{aligned}
\label{eq:trilevel_level_summary}
\end{equation}

\textbf{Optimization Schema.} Solving the aforementioned objective involves a computationally expensive tri-level optimization problem, which requires multi-level optimal. To reduce the computational cost, we sequentially optimize the surrogate model $f_c$ and the trigger generator $f_g$ for several epochs instead of optimizing to convergence, at each update of the condensed graph $\mathcal{S}$. This significantly reduces the number of iterations, improving the efficiencies.

\textit{Surrogate Model.} To further mitigate the computational complexity, we follow~\cite{adversarialGNNs_2019iclr,dai2023www-unnoticeable} to update surrogate model ${\theta}_c$ for $T$ iterations with fixed generator ${\theta}_g$ and condensed graph $\mathcal{S}$ to approximate ${\theta}_c^*$:
\begin{equation}
    \label{eq:update-surrogate}
    {\theta}_c^{t+1}={\theta}_c^t-\alpha_c\nabla_{{\theta}_c}\mathcal{L}_c({\theta}_c,\mathcal{S},{\theta}_g),
\end{equation}
where $\alpha_c$ is the learning rate for surrogate model training and ${\theta}_c^t$ denotes the model parameters after $t$ iterations.

\textit{Trigger Generator.} We also apply the $M$-iterations approximation to the optimization for trigger generator ${\theta}_g$ with updated ${\theta}_c^T$ and the fixed $\mathcal{S}$:
\begin{equation}
    \label{eq:update-trigger-gen}
    {\theta}_g^{m+1} = {\theta}_g^{m}-\alpha_g\nabla_{{\theta}_g}\mathcal{L}_g({\theta}_c^T,\mathcal{S},{\theta}_g),
\end{equation}
where $\alpha_g$ is the learning rate for the trigger generator.

\textit{Condensed Graph.} In the outer iteration, we compute the gradients of the condensed graph $\mathcal{S}$ based on the trigger generator ${\theta}_g^M$ and the trained surrogate model ${\theta}_c^T$ as following:
\begin{equation}
    \label{eq:update-condensed-graph}
    \mathcal{S}^{k+1} = \mathcal{S}^k-\alpha_{\mathcal{S}}\nabla_{\mathcal{S}}\mathcal{L}_s({\theta}_c^T,\mathcal{S},{\theta}_g^M),
\end{equation}
where $\alpha_{\mathcal{S}}$ is the learning rate of updating condensed graph. The summarization of the whole BGC's optimization process could be found in Algorithm~\ref{algorithm:BGC}.

{
\textbf{Convergence Property.} Follow the proof of Theorem 1 and Theorem 2 in Doscond~\cite{doscond-kdd2022}, we adopt SGC as surrogate model, i.e., $f_{\theta_c}(\mathcal{G}^P|\theta_c)=\mathbf{A}^P\mathbf{X}^P\mathbf{W}$, where $\mathbf{A}^P\in\{0,1\}^{N^P\times N^P}$ is the adjacency matrix of poisoned graph, $\mathbf{X}^P\in \mathcal{R}^{N^P\times d}$ denotes the feature matrix for the poisoned graph and $\theta_c=\mathbf{W}$. A pivotal design decision involves ensuring full connectivity among the trigger nodes generated by the trigger generator $f_{\theta_g}$, thereby maintaining the invariance of $\mathbf{A}^P$,  throughout the condensation process. Consequently, as illustrated in Algorithm~\ref{algorithm:BGC}, the parameters $\mathbf{W}^P$ ($\theta_c$) and $\mathbf{X}^P$ ($\theta_g$) are updated sequentially. This allows us to conceptualize them as components of a unified optimization task, denoted by $\hat{\mathbf{W}}=\mathbf{W}^P\mathbf{X}^P$ and $\theta_{cg}$, which combines $\theta_c$ and $\theta_g$. This reformulation enables the transformation of the tri-level optimization problem into a bi-level optimization framework. In this framework, the inner optimization focuses on refining the “GNN model” parameterized by $\theta_{cg}$ over $M+T$ iterations, while the outer optimization is dedicated to optimize the condensed graph $\mathcal{S}$. Then, we can transform the tri-level optimization problem in Eq.(\ref{eq:trilevel_level_summary}) into following bi-levle problem (graph condensation):
$$
    \min_{\mathcal{S}} \;\mathcal{L}_s(\mathcal{S},{\theta}^*_{cg})\;\textit{ s.t. } \; 
    {{\theta}^*_{cg}} =  \arg\min_{\theta_{cg}}\; \mathcal{L}_g(\mathcal{S},{\theta}_{cg}). \;\\
$$

While we focus on node classification task in this work, according to Theorem 2 in Doscond~\cite{doscond-kdd2022}, we have: $\mathcal{L}_s(\mathcal{S},{\theta}_{cg})$ is convex and $L$-smooth. We assume that $\mathbb{E}||\nabla_{\mathcal{S}}\mathcal{L}_s(\mathcal{S},{\theta}^*_{cg})-\nabla_\mathcal{S}\mathcal{L}_s(\mathcal{S},{\theta}^{M+T}_{cg})||^2\leq \sigma^2$ and stepsize $\eta<\frac{1}{L}$. We denote $\mathcal{S}$'s gradient mapping at $k$-th iteration by 
    $\mathcal{G}^k=\frac{1}{\eta}\mathcal{L}_s(\mathcal{S},{\theta}^{M+T}_{cg}).$

According to the proof for Theorem 1 in DConRec~\cite{jiahao2023DConRec}, we can have following conclusion given above assumptions:
$$\frac{1}{K} \sum_{t=1}^K \mathbb{E}\left\|\mathcal{G}^t\right\|^2 \leq \frac{8-2 L \eta}{2-L \eta} \sigma^2,$$ 
when $K\rightarrow \infty$. $K$ is the total iterations of $\mathcal{S}$. Therefore, BGC is provably convergent.
}
\input{sections/experiment-results/compare-with-baselines/main}
\section{Experimental Settings}
\textbf{Datasets.} We evaluate our proposed attack on two transductive datasets, i.e., Cora, Citeseer~\cite{gcn2017iclr}, and two inductive datasets, i.e., Flickr~\cite{graphsaint2020iclr}, Reddit~\cite{reddit2017nips}. All the datasets have public splits and we download them from Pytorch Geometric~\cite{pytorch2019}, following those splits throughout the experiments. The statistics of the datasets and splits are summarized in Table~\ref{tab:data-statistic}.

\textbf{Graph Condensation Methods.} In this paper, we incorporate four  prevalent graph condensation methods to test the attack performance of our proposed method: 1) DC-Graph~\cite{zhaoICLR2021DC,iclr2022gcond} is the graph-based variant of the general dataset condensation method DC~\cite{zhaoICLR2021DC}, 2) GCond~\cite{iclr2022gcond} is one representative graph condensation method, 3) GCond-X~\cite{iclr2022gcond} is the variant of GCond that discards the structure information of condensed graph for GNNs' training and 4) GC-SNTK~\cite{wangling2023fastGC} reforms the graph condensation as a Kernel Ridge Regression (KRR) task and it is based on the Structure-based Neural Tangent Kernel (SNTK). While DC-graph, GCond, and GCond-X offer flexibility in utilizing different GNNs for the condensation and test stages, we default to adopting the best-performing of setting SGC~\cite{SGC2019ICML} as the backbone for condensation and GCN~\cite{gcn2017iclr} for testing. For the parameter settings of DC-graph, GCond, and GCond-X, we follow the settings described in~\cite{iclr2022gcond}. Regarding the GC-SNTK method, we adopt the settings described in the original paper~\cite{wangling2023fastGC}.

\textbf{Evaluation Metrics.} To evaluate the effectiveness, we adopt \textit{attack success rate (ASR)} and \textit{clean test accuracy (CTA)} as the metrics. The \textit{ASR} measures the attack effectiveness of the backdoored GNN on the triggered testing dataset and the \textit{CTA} measures the utility of the backdoored GNN on the clean testing dataset. Both \textit{ASR} and \textit{CTA} range from $0.0$ to $1.0$. The higher value of \textit{ASR} denotes the better attack performance. The closer the \textit{CTA} of the backdoored GNN to the one of a clean GNN,  the better the backdoored GNN’s utility is. 

\textbf{Targeted GNNs.} For DC-graph, GCond, and GCond-X, we default to adopting GCN as the testing architecture as described previously. Since GC-SNTK is based on a neural tangent kernel (NTK), it is only applicable to the NTK-based model. To validate the generalization of our proposed attack method, we also utilize the condensed graph to backdoor other architectures of GNN regarding DC-graph, GCond, and GCond-X methods, presented in section~\ref{subsec:cross-architect}. 

\textbf{Runtime Configuration.} We have examined the attack performance of our method under these condensation ratios: $\{1.30\%,2.60\%,5.20\%,\}$ for Cora, $\{0.90\%,1.80\%,3.60\%\}$ for Citeseer, $\{0.10\%,0.50\%,1.00\%\}$ for Flickr and $\{0.05\%,0.10\%,0.20\%\}$ for Reddit. The poisoning ratio of Cora and Citeseer defaults to $0.1$. The poisoning number of Flickr defaults to $80$ and Reddit's defaults to $180$. All the experiments are repeated 3 times. For each run, we follow the same experimental setup
laid out before. We report the mean and standard deviation
of each metric to evaluate the attack performance.

\textbf{Implementation Details.} Regarding the learning rates, optimizer, epochs for condensation, and model training, we all follow the settings described in the original papers of the condensation methods~\cite{iclr2022gcond,zhaoICLR2021DC,wangling2023fastGC}. Due to the design of BGC, the iteration number of the optimizing trigger generator is the same as the one of condensation (i.e., $1000$). The trigger generator's learning rate is searched within $\{0.01,0.05,0.1,0.5\}$ and its optimizer is Adam. The trigger size set to 4 by default, following the setting in~\cite{dai2023www-unnoticeable,icml2019evasionGraphAttack,aaai2020evasionGraphAttack,gta2021usenix}.

\section{Experiment}
In this section, we present the performance of our method against graph condensation. The extensive experiments are devised to answer the following questions:
\begin{itemize}[leftmargin=*]
    \item \textbf{RQ1}: Can BGC achieve high attack performance and preserve the model utility of GNNs?
    \item \textbf{RQ2}: Can the condensed graph by BGC generalize to successfully attack different architectures of GNNs?
    \item \textbf{RQ3}: How does BGC perform against defense methods?
\end{itemize}
Besides, we also conduct various studies to analyze the properties of BGC, investigating how different settings and hyper-parameters (e.g., ablation study on poisoned node selection, size of triggers, number of GNNs' layers, condensation epochs, and poison ratios) affect the attack performance and the condensed graph utility. 
\subsection{Attack Performance and Model Utility \textbf{(RQ1)}}
\textbf{Attack Performance.} 
To measure BGC's attack performance, we conduct a comparative evaluation of the ASR score between the backdoored GNN and the clean GNN, which are reported as {ASR} and {C-ASR} in Table~\ref{tab:model-utility-attack-performance}, respectively. We can observe that the attack on clean model achieves low attack success scores {C-ASR} (i.e., ranging below $18\%$). In contrast, all of the {ASR} scores achieved by our method are over $95.0\%$. For example, the {ASR} scores on the datasets Cora and Citeseer are $100\%$ with condensation methods DC-Graph, GCond, and GCond-X. Besides, the {ASR} scores are also much higher than the scores of clean GNNs {C-ASR}. This indicates the decent attack performance of our proposed method. 

\textbf{Model Utility.} To measure the utility of the backdoored GNNs, i.e., evaluating whether our attacks undermines the performance of GNNs on the primary task. For a successful backdoor attack, the backdoored GNN should be as good as the clean GNN on the task, given clean test data. As observed from Table~\ref{tab:model-utility-attack-performance}, the {CTA} scores of the backdoored GNN are close to the {C-CTA} scores of the clean GNN. For example, the value of {C-CTA} for the Cora dataset is $75.17$ with the setting that the condensation method is DC-Graph and the condensation ratio is $1.30\%$. In the same case, the {CTA} of the backdoored GNN is $75.90$. Besides, the largest gap between {C-CTA} and {CTA} is the case that the condensation method is GCond-X, and the condensation ratio is $0.50\%$ for the dataset Flickr, where the {CTA} ($45.21\%$) drops by $3.15\%$ compared to {C-CTA} ($46.68\%$). This side effect is within the acceptable performance, indicating utility preservation of our method.

\begin{table}[]
\centering
\caption{Study on Different GNN Architectures. Our method achieves high ASR and CTA scores close to C-CTA across GNN architectures.}
\vskip -0.08in
\resizebox{0.96\linewidth}{!}{ 
\begin{tabular}{|c|c|c|c|c|c|}
\hline
GNN            & Metrics  & Cora & Citeseer & Flickr & Reddit \\ \hline
\multirow{3}{*}{GCN} 
    & \cellcolor{gray!20}C-CTA & \cellcolor{gray!20}81.27 (0.33) & \cellcolor{gray!20}71.43 (0.33) & \cellcolor{gray!20}46.91 (0.41)  & \cellcolor{gray!20}89.20 (0.13) \\
    & \cellcolor{gray!20}CTA   & \cellcolor{gray!20}80.67 (0.52) & \cellcolor{gray!20}71.57 (1.32) & \cellcolor{gray!20}46.84 (0.09)  & \cellcolor{gray!20}90.37 (0.22) \\
    & ASR                      & 100.0 (0.00) & 100.0 (0.00) & 99.77 (0.06)  & 99.99 (0.01) \\ \hline
\multirow{3}{*}{SAGE} 
    & \cellcolor{gray!20}C-CTA & \cellcolor{gray!20}79.73 (0.66) & \cellcolor{gray!20}73.03 (0.05) & \cellcolor{gray!20}46.56 (0.08)  & \cellcolor{gray!20}90.25 (0.19) \\
    & \cellcolor{gray!20}CTA   & \cellcolor{gray!20}77.03 (0.53) & \cellcolor{gray!20}70.23 (1.52) & \cellcolor{gray!20}46.99 (0.08)  & \cellcolor{gray!20}88.33 (0.92) \\
    & ASR                      & 100.0 (0.00) & 100.0 (0.00) & 97.28 (1.79)  & 99.85 (0.12) \\ \hline
\multirow{3}{*}{SGC} 
    & \cellcolor{gray!20}C-CTA & \cellcolor{gray!20}78.97 (0.33) & \cellcolor{gray!20}72.07 (0.20) & \cellcolor{gray!20}46.50 (0.37)  & \cellcolor{gray!20}90.99 (0.22) \\
    & \cellcolor{gray!20}CTA   & \cellcolor{gray!20}80.23 (0.21) & \cellcolor{gray!20}68.73 (1.26) & \cellcolor{gray!20}47.16 (0.02)  & \cellcolor{gray!20}90.65 (0.13) \\
    & ASR                      & 100.0 (0.00) & 100.0 (0.00) & 92.14 (0.58)  & 99.99 (0.00) \\ \hline
\multirow{3}{*}{MLP} 
    & \cellcolor{gray!20}C-CTA & \cellcolor{gray!20}78.77 (0.88) & \cellcolor{gray!20}70.67 (0.46) & \cellcolor{gray!20}42.24 (0.30)  & \cellcolor{gray!20}43.66 (1.04) \\
    & \cellcolor{gray!20}CTA   & \cellcolor{gray!20}76.20 (0.86) & \cellcolor{gray!20}53.57 (7.69) & \cellcolor{gray!20}46.60 (0.25)  & \cellcolor{gray!20}42.75 (0.95) \\
    & ASR                      & 100.0 (0.00) & 100.0 (0.00) & 95.45 (2.28)  & 100.0 (0.00) \\ \hline
\multirow{3}{*}{APPNP} 
    & \cellcolor{gray!20}C-CTA & \cellcolor{gray!20}79.10 (0.29) & \cellcolor{gray!20}71.30 (0.22) & \cellcolor{gray!20}45.90 (0.30)  & \cellcolor{gray!20}88.53 (0.44) \\
    & \cellcolor{gray!20}CTA   & \cellcolor{gray!20}79.23 (0.60) & \cellcolor{gray!20}47.67 (0.39) & \cellcolor{gray!20}46.74 (0.07)  & \cellcolor{gray!20}88.41 (0.02) \\
    & ASR                      & 100.0 (0.00) & 100.0 (0.00) & 100.0 (0.00)  & 100.0 (0.00) \\ \hline
\multirow{3}{*}{Cheby.} 
    & \cellcolor{gray!20}C-CTA & \cellcolor{gray!20}78.23 (0.60) & \cellcolor{gray!20}66.90 (1.13) & \cellcolor{gray!20}42.37 (0.01)  & \cellcolor{gray!20}75.79 (0.73) \\
    & \cellcolor{gray!20}CTA   & \cellcolor{gray!20}77.37 (1.02) & \cellcolor{gray!20}66.33 (0.62) & \cellcolor{gray!20}41.33 (1.30)  & \cellcolor{gray!20}72.78 (1.21) \\
    & ASR                      & 100.0 (0.00) & 100.0 (0.00) & 88.63 (3.28)  & 98.54 (1.71) \\ \hline
\end{tabular}
}
\label{tab:cross-architectures}
\vskip -0.25in
\end{table}

\subsection{Attack Performance Comparison}
\label{sec:attack-perform-compare}
To further validate the effectiveness of our design, we compare BGC with previous {\textbf{backdoor attack baselines}: 1) GTA~\cite{gta2021usenix} is the most representative backdoor attack method on the graph, which injects triggers to the graph during model training, and 2) DOORPING~\cite{ndss2023doorping} is a backdoor attack against dataset distillation for image data, which directly learns triggers for the images.} To adapt GTA in the context of graph condensation, we apply it to the original graph and then utilize the poisoned graph for condensation. To transform DOORPING for graph data, we learn the universal triggers for all the nodes following training procedures described in the original paper~\cite{ndss2023doorping}. Besides, regarding the poisoned nodes, we utilize the selection module of our proposed method. The experimental results
are presented in Figure~\ref{fig:compare-with-baselines}. Although GTA and DOORPING can perform well in some cases, they are still inferior to BGC, and they fail to launch the effective attack in many cases (i.e., for all the settings with the condensation method GC-SNTK, the attack success rate of these two methods is less than $88.81\%$). Besides, their attacks can lead to a significant drop in the GNN's utility. For instance, under the setting that the condensation ratio is $0.90\%$ and the condensation method is GCond-X for dataset Citeseer, the CTAs of GTA, and DOOPRPING are $69.00$ and $55.47$, dropping by $6.45\%$ and $24.80\%$ respectively. \textbf{This demonstrates the superiority of our method and reveals that previous methods cannot launch a successful backdoor attack against graph condensation.}  
\subsection{Study on Different GNN Architectures \textbf{(RQ2)}}
\label{subsec:cross-architect}
Since we envision the attacker as the graph condensation service provider, the attacker does not know which GNN the customer will train with the condensed graph. Therefore, it is essential to test the effectiveness of BGC in backdooring different architectures of GNNs. Specifically, we utilize the graphs condensed by BGC with condensation method GCond for the training of various architectures of GNNs: GCN~\cite{gcn2017iclr}, GraphSage~\cite{reddit2017nips}, SGC~\cite{SGC2019ICML}, MLP~\cite{hu2021graphmlp}, APPNP~\cite{iclr2019appnp} and ChebyNet~\cite{chebynet2016nips}. The condensation ratios for four datasets are: Cora, $2.60\%$; Citeseer, $0.90\%$; Flickr, $1.00\%$; Reddit, $0.10\%$. The results are reported in Table~\ref{tab:cross-architectures}. We can observe that the {ASR} scores in many cases are $100\%$ (i.e., for dataest Cora and Citeseer) and the rest are over $92\%$, except the one for ChebyNet on dataset Flickr. Besides, the {CTA} scores in most cases are close to the {C-CTA}. This demonstrates that \textbf{the condensed graphs by our proposed method can carry out effective backdoor attacks against different GNN models.}
\begin{figure}
    \centering
    \setlength\tabcolsep{2pt}
    \begin{tabular}{c}
        \resizebox{0.45\linewidth}{!}{
            \begin{tikzpicture}
    \begin{customlegend}[
        legend columns=2,
        legend style={
            anchor=north,             
            draw=none,               
            fill=white,               
            text depth=0pt,           
            font=\footnotesize,       
            /tikz/every even column/.append style={column sep=0.3cm},
            legend image code/.code={
                \draw[fill=##1, draw=white, /tikz/.cd,yshift=-0.3em]
                (0cm,0cm) rectangle (0.35em,0.5em); 
            }
        },
        legend entries={
            \textsc{$\text{BGC}_{\text{Rand}}$},
            \textsc{$\text{BGC}$},
        }
        ] 
        \addlegendimage{ybar,ybar legend, 
        fill={rgb, 255:red,100;green,130;blue,180},
        draw=black}
        \addlegendimage{ybar,ybar legend, 
        fill={rgb, 255:red,127;green,187;blue,209},
        draw=black}

        \end{customlegend}
\end{tikzpicture}
        }\\
        \begin{tabular}{cc}
            \begin{subfigure}{0.49\linewidth}
                \resizebox{1.0\linewidth}{!}{
\scalebox{1.0}{
\begin{tikzpicture}
\begin{axis}[
    ybar,
    enlargelimits=0.25, 
    enlarge y limits={upper,value=0.05},  
    ylabel={CTA (\%)},
    xlabel style={font=\huge},  
    ylabel style={font=\huge},  
    symbolic x coords={0.10\%, 0.50\%, 1.00\%},
    xtick=data,
    ymin=41, ymax=47.5,
    ytick={41, 43, 45, 47},
    xticklabel style={font=\huge, rotate=0},
    yticklabel style={font=\huge},  
    grid=none,  
    ymajorgrids=true,  
    axis background/.style={
        fill={rgb, 255:red,228;green,228;blue,237},
        fill opacity=0.7  
    },
    grid style={color=white},  
    axis line style={color=white},  
    tick style={color=white},  
    bar width=17pt,
    yticklabel style={
    /pgf/number format/.cd,
    fixed,
    fixed zerofill,
    precision=0  
    },
    xtick pos=bottom,  
    every axis plot/.append style={
        fill,
        draw=none
    },
]
    \addplot+[
        fill={rgb, 255:red,100;green,130;blue,180},
        draw=black,
        error bars/.cd,
        y dir=both,
        y explicit,
        error bar style={draw=black, line width=1pt},  
        error mark options={rotate=90, mark size=2pt, draw=black}  
    ] coordinates {
        (0.10\%,45.94) +- (0,0.31)
        (0.50\%,45.39) +- (0,0.52)
        (1.00\%,46.04) +- (0,0.67)
    };

    \addplot+[
        fill={rgb, 255:red,127;green,187;blue,209},  
        draw=black,
        error bars/.cd,
        y dir=both,
        y explicit,
        error bar style={draw=black, line width=1pt},  
        error mark options={rotate=90, mark size=2pt, draw=black}  
    ] coordinates {
        (0.10\%,46.48) +- (0,0.21)
        (0.50\%,46.44) +- (0,0.13)
        (1.00\%,46.77) +- (0,0.30)
    };
\end{axis}
\end{tikzpicture}
}

  
                }
            \end{subfigure} 
            &
            \begin{subfigure}{0.49\linewidth}
                \resizebox{1.0\linewidth}{!}{
\scalebox{1.0}{
\begin{tikzpicture}
\begin{axis}[
    ybar,
    enlargelimits=0.25, 
    enlarge y limits={upper,value=0.05},  
    ylabel={CTA (\%)},
    ylabel style={font=\huge},  
    symbolic x coords={0.05\%, 0.10\%, 0.20\%},
    xtick=data,
    ymin=73.5, ymax=91.5,
    ytick={75, 80, 85, 90},
    xticklabel style={font=\huge, rotate=0},
    yticklabel style={font=\huge},  
    grid=none,  
    ymajorgrids=true,  
    axis background/.style={
        fill={rgb, 255:red,228;green,228;blue,237},
        fill opacity=0.7  
    },
    grid style={color=white},  
    axis line style={color=white},  
    tick style={color=white},  
    bar width=17pt,
    xtick pos=bottom,  
    every axis plot/.append style={
        fill,
        draw=none
    },
]
    \addplot+[
        fill={rgb, 255:red,100;green,130;blue,180},
        draw=black,
        error bars/.cd,
        y dir=both,
        y explicit,
        error bar style={draw=black, line width=1pt},  
        error mark options={rotate=90, mark size=2pt, draw=black}  
    ] coordinates {
        (0.05\%,85.05) +- (0,0.64)
        (0.10\%,88.22) +- (0,0.45)
        (0.20\%,89.44) +- (0,0.10)
    };

    \addplot+[
        fill={rgb, 255:red,127;green,187;blue,209}, 
        draw=black,
        error bars/.cd,
        y dir=both,
        y explicit,
        error bar style={draw=black, line width=1pt},  
        error mark options={rotate=90, mark size=2pt, draw=black}  
    ] coordinates {
        (0.05\%,85.38) +- (0,0.42)
        (0.10\%,89.14) +- (0,0.05)
        (0.20\%,90.38) +- (0,0.42)
    };
\end{axis}
\end{tikzpicture}
}
                }
            \end{subfigure} 
            \\
            \begin{subfigure}{0.483\linewidth}
                \resizebox{1.0\linewidth}{!}{
\scalebox{1.0}{
\begin{tikzpicture}
\begin{axis}[
    ybar,
    enlargelimits=0.25, 
    enlarge y limits={upper,value=0.05},  
    ylabel={ASR (\%)},
    ylabel style={font=\huge},  
    xlabel={Cond. Ratio of Flickr},
    xlabel style={font=\huge},  
    symbolic x coords={0.10\%, 0.50\%, 1.00\%},
    xtick=data,
    xticklabel style={font=\huge, rotate=0},
    yticklabel style={font=\huge},  
    grid=none,  
    ymajorgrids=true,  
    axis background/.style={
        fill={rgb, 255:red,228;green,228;blue,237},
        fill opacity=0.7  
    },
    grid style={color=white},  
    axis line style={color=white},  
    tick style={color=white},  
    bar width=17pt,
    ymin=93, ymax=100.5,  
    xtick pos=bottom,  
    every axis plot/.append style={
        fill,
        draw=none
    },
]

    \addplot+[
        fill={rgb, 255:red,100;green,130;blue,180},
        draw=black,
        error bars/.cd,
        y dir=both,
        y explicit,
        error bar style={draw=black, line width=1pt},  
        error mark options={rotate=90, mark size=2pt, draw=black}  
    ] coordinates {
        (0.10\%,96.98) +- (0,0.02)
        (0.50\%,96.87) +- (0,1.25)
        (1.00\%,98.79) +- (0,0.80)
    };

    \addplot+[
        fill={rgb, 255:red,127;green,187;blue,209}, 
        draw=black,
        error bars/.cd,
        y dir=both,
        y explicit,
        error bar style={draw=black, line width=1pt},  
        error mark options={rotate=90, mark size=2pt, draw=black}  
    ] coordinates {
        (0.10\%,99.98) +- (0,0.02)
        (0.50\%,99.25) +- (0,0.65)
        (1.00\%,99.11) +- (0,0.40)
    };
    
\end{axis}
\end{tikzpicture}
}
                }
            \end{subfigure}
            &
            \begin{subfigure}{0.49\linewidth}
                \resizebox{1.0\linewidth}{!}{
\scalebox{1.0}{
\begin{tikzpicture}
\begin{axis}[
    ybar,
    enlargelimits=0.25, 
    enlarge y limits={upper,value=0.05},  
    ylabel={ASR (\%)},
    ylabel style={font=\huge},  
    xlabel={Cond. Ratio of Reddit},
    xlabel style={font=\huge},  
    symbolic x coords={0.05\%, 0.10\%, 0.20\%},
    xtick=data,
    xticklabel style={font=\huge, rotate=0},
    yticklabel style={font=\huge},  
    grid=none,  
    ymajorgrids=true,  
    axis background/.style={
        fill={rgb, 255:red,228;green,228;blue,237},
        fill opacity=0.7  
    },
    grid style={color=white},  
    axis line style={color=white},  
    tick style={color=white},  
    bar width=17pt,
    ymin=93, ymax=100.5,  
    xtick pos=bottom,  
    every axis plot/.append style={
        fill,
        draw=none
    },
]

    \addplot+[
        fill={rgb, 255:red,100;green,130;blue,180},
        draw=black,
        error bars/.cd,
        y dir=both,
        y explicit,
        error bar style={draw=black, line width=1pt},  
        error mark options={rotate=90, mark size=2pt, draw=black}  
    ] coordinates {
        (0.05\%,97.37) +- (0,1.09)
        (0.10\%,94.54) +- (0,0.46)
        (0.20\%,98.79) +- (0,0.80)
    };

    \addplot+[
        fill={rgb, 255:red,127;green,187;blue,209}, 
        draw=black,
        error bars/.cd,
        y dir=both,
        y explicit,
        error bar style={draw=black, line width=1pt},  
        error mark options={rotate=90, mark size=2pt, draw=black}  
    ] coordinates {
        (0.05\%,99.90) +- (0,0.03)
        (0.10\%,99.93) +- (0,0.02)
        (0.20\%,99.90) +- (0,0.05)
    };
    
\end{axis}
\end{tikzpicture}
}
                }
            \end{subfigure}
        \end{tabular}
\\
    \end{tabular}
    \vskip -0.05in
    \caption{Ablation Study on Poisoned Node Selection.}
    \label{fig:main-random-select-ablation}
    \vskip -0.25in
\end{figure}
\subsection{Defenses \textbf{(RQ3)}}
\begin{table*}[]
\centering
\vskip -0.15in
\caption{Attack Performance against Defenses. `\textit{OOM}' denotes out of memory and `\textit{--}' denotes not applicable. {The defenses} (\textit{Prune}, \textit{Randsmooth})  {suffer from a trade-off, with significant CTA losses ($\Delta$ CTA) for limited ASR reduction} ($\Delta$ ASR).}
\vskip -0.05in
\resizebox{0.9\linewidth}{!}{ 
\begin{threeparttable}
\newcolumntype{G}{>{\columncolor{gray!20}}c} 
\begin{tabular}{|c|c|c|cGcG|cGcG|cc|}
\hline
 &
   &
   &
  \multicolumn{4}{c|}{Defense-Prune} &
  \multicolumn{4}{c|}{Defense-Randsmooth} &
  \multicolumn{2}{c|}{Backdoor} \\ \cline{4-13} 
\multirow{-2}{*}{\begin{tabular}{@{}c@{}} Cond. \\ Method \end{tabular}} &
  \multirow{-2}{*}{Datasets} &
  \multirow{-2}{*}{Ratio (r)} &
  CTA &
  $\Delta$ CTA &
  ASR &
  $\Delta$ ASR &
  CTA &
  $\Delta$ CTA &
  ASR &
  $\Delta$ ASR &
  CTA &
  ASR \\ \hline
 &
   &
  1.30\% &
  45.40 &
  {-44.11\%} &
  75.60 &
  {-24.40\%} &
  78.70 &
  {-3.11\%} &
  99.07 &
  {-0.93\%} &
  81.23 &
  100.0 \\
 &
   &
  2.60\% &
  54.67 &
  {-32.23\%} &
  65.10 &
  {-34.90\%} &
  78.67 &
  {-2.48\%} &
  99.10 &
  {-0.90\%} &
  80.67 &
  100.0 \\
 &
  \multirow{-3}{*}{Cora} &
  5.20\% &
  55.97 &
  {-30.64\%} &
  99.00 &
  {-1.00\%} &
  80.10 &
  {-0.74\%} &
  98.83 &
  {-1.17\%} &
  80.70 &
  100.0 \\ \cline{2-13} 
 &
   &
  0.90\% &
  36.17 &
  {-49.46\%} &
  95.10 &
  {-4.90\%} &
  68.86 &
  {-3.79\%} &
  99.30 &
  {-0.70\%} &
  71.57 &
  100.0 \\
 &
   &
  1.80\% &
  48.87 &
  {-31.20\%} &
  85.70 &
  {-14.30\%} &
  62.90 &
  {-11.45\%} &
  99.37 &
  {-0.63\%} &
  71.03 &
  100.0 \\
 &
  \multirow{-3}{*}{Citeseer} &
  3.60\% &
  40.43 &
  {-42.73\%} &
  71.70 &
  {-28.30\%} &
  60.80 &
  {-13.88\%} &
  98.70 &
  {-1.30\%} &
  70.60 &
  100.0 \\ \cline{2-13} 
 &
   &
  0.10\% &
  41.52 &
  {-10.79\%} &
  90.03 &
  {-9.82\%} &
  45.57 &
  {-2.08\%} &
  97.53 &
  {-2.30\%} &
  46.54 &
  99.83 \\
 &
   &
  0.50\% &
  40.91 &
  {-13.23\%} &
  89.24 &
  {-10.73\%} &
  45.12 &
  {-4.31\%} &
  88.04 &
  {-11.93\%} &
  47.15 &
  99.97 \\
 &
  \multirow{-3}{*}{Flickr} &
  1.00\% &
  42.31 &
  {-9.67\%} &
  88.33 &
  {-11.47\%} &
  45.93 &
  {-1.94\%} &
  94.22 &
  {-5.56\%} &
  46.84 &
  99.77 \\ \cline{2-13} 
 &
   &
  0.05\% &
  67.14 &
  {-24.14\%} &
  91.28 &
  {-8.57\%} &
  72.99 &
  {-17.53\%} &
  99.97 &
  {0.13\%} &
  88.5 &
  99.84 \\
 &
   &
  0.10\% &
  40.49 &
  {-55.20\%} &
  87.86 &
  {-12.13\%} &
  87.14 &
  {-3.57\%} &
  99.95 &
  {-0.04\%} &
  90.37 &
  99.99 \\
\multirow{-12}{*}{GCond} &
  \multirow{-3}{*}{Reddit} &
  0.20\% &
  56.74 &
  {-37.23\%} &
  88.15 &
  {-11.01\%} &
  87.01 &
  {-3.75\%} &
  95.73 &
  {-3.36\%} &
  90.40 &
  99.06 \\ \hline\hline
 &
   &
  1.30\% &
  72.03 &
  {-5.60\%} &
  73.67 &
  {-26.33\%} &
  75.70 &
  {-0.79\%} &
  98.87 &
  {-1.13\%} &
  76.30 &
  100.0 \\
 &
   &
  2.60\% &
  71.63 &
  {-8.28\%} &
  67.73 &
  {-32.27\%} &
  75.47 &
  {-3.37\%} &
  98.87 &
  {-1.13\%} &
  78.10 &
  100.0 \\
 &
  \multirow{-3}{*}{Cora} &
  5.20\% &
  77.53 &
  {-2.36\%} &
  89.80 &
  {-10.20\%} &
  76.93 &
  {-3.11\%} &
  98.90 &
  {-1.10\%} &
  79.40 &
  100.0 \\ \cline{2-13} 
 &
   &
  0.90\% &
  56.13 &
  {-23.14\%} &
  81.39 &
  {-18.61\%} &
  56.60 &
  {-22.50\%} &
  88.24 &
  {-11.76\%} &
  73.03 &
  100.0 \\
 &
   &
  1.80\% &
  65.37 &
  {-9.71\%} &
  80.10 &
  {-19.90\%} &
  55.43 &
  {-23.44\%} &
  88.70 &
  {-11.30\%} &
  72.40 &
  100.0 \\
 &
  \multirow{-3}{*}{Citeseer} &
  3.60\% &
  60.10 &
  {-16.68\%} &
  87.20 &
  {-12.80\%} &
  55.10 &
  {-23.61\%} &
  81.38 &
  {-18.62\%} &
  72.13 &
  100.0 \\ \cline{2-13} 
 &
   &
  0.10\% &
  42.14 &
  {-8.69\%} &
  65.93 &
  {-32.90\%} &
  45.02 &
  {-2.45\%} &
  99.97 &
  {1.74\%} &
  46.15 &
  98.26 \\
 &
   &
  0.50\% &
  43.99 &
  {-2.70\%} &
  70.06 &
  {-29.64\%} &
  45.83 &
  {1.37\%} &
  93.65 &
  {-5.96\%} &
  45.21 &
  99.58 \\
 &
  \multirow{-3}{*}{Flickr} &
  1.00\% &
  45.11 &
  {-1.12\%} &
  80.07 &
  {-16.17\%} &
  45.21 &
  {-0.90\%} &
  98.29 &
  {2.91\%} &
  45.62 &
  95.51 \\ \cline{2-13} 
 &
   &
  0.05\% &
  87.12 &
  {-0.40\%} &
  90.15 &
  {-9.75\%} &
  88.26 &
  {0.90\%} &
  98.87 &
  {-1.02\%} &
  87.47 &
  99.89 \\
 &
   &
  0.10\% &
  88.69 &
  {-0.50\%} &
  92.78 &
  {-7.02\%} &
  89.55 &
  {0.46\%} &
  94.15 &
  {-5.64\%} &
  89.14 &
  99.78 \\
\multirow{-12}{*}{GCond-X} &
  \multirow{-3}{*}{Reddit} &
  0.20\% &
  89.80 &
  {-0.32\%} &
  91.72 &
  {-6.02\%} &
  89.90 &
  {-0.21\%} &
  98.64 &
  {1.07\%} &
  90.09 &
  97.6 \\ \hline
\end{tabular}
\end{threeparttable}
}
\label{tab:performance-under-defense}
\vskip -0.2in
\end{table*}

To mitigate the threats of backdoor attacks, various defense methods have been proposed. In this section, we evaluate the robustness of our proposed method against two representative graph defense methods: 1) Prune~\cite{dai2023www-unnoticeable} is a dataset-level defense, which prunes edges linking nodes with low cosine similarity in the condensed graphs, and 2) Randsmooth~\cite{sacmat2021BaGNN} is a model-level defense, which randomly subsamples $d$ subgraphs to generate $d$ outputs, using a voting mechanism for the final prediction. The results are presented in Table~\ref{tab:performance-under-defense}, where we report the {CTA} and {ASR} scores of BGC, the scores of BGC under the defenses and the decreased ratio of those scores. By analyzing the score changes, we assess the effectiveness of the defense mechanisms in mitigating our proposed attack.

In our experiment, we implement Prune by removing the edges connecting nodes that fall within the lowest $20\%$ of cosine similarities. As we can observe from the table, most {ASR} scores decrease. However, the {CTA} scores also drop significantly, often more than the ASR scores. For example, when the condensation ratio is $3.60\%$ for the Citeseer dataset using the GCond method, the {CTA} score drops by $42.73\%$, while the ASR score decreases by only $28.30\%$. In conclusion, \textbf{Prune cannot defend our proposed BGC while it severely suffers from a utility-defense trade-off.} 
Regarding Randsmooth, we implement it by sampling different sub-structures for the propagations in different layers. From Table~\ref{tab:performance-under-defense}, we can observe that the decrease in {ASR} scores caused by the defense is limited (i.e., less than $2\%$ in most cases) while the decrease of {CTA} scores is even larger in most cases.
A significant decline in {ASR} generally leads to a substantial decline in {CTA}. For instance, on dataset Citeseer with GCond-X, although the {ASR} scores drop by around $12\%$ and $19\%$, the {CTA} scores drop by over $22\%$. This indicates that \textbf{Randsmooth also suffers from the utility-defense trade-off and has limited defensive effectiveness.}
\begin{figure*}
    \vskip -0.15in
    \centering
    \setlength\tabcolsep{1.5pt}
    \begin{tabular}{cccc}
    \begin{subfigure}{0.25\linewidth}
    \centering
            \resizebox{1.0\linewidth}{!}{
                \begin{tikzpicture}
\begin{axis}[
    xlabel={Epoch},
    ylabel={ASR ($\%$)},
    ylabel style={font=\huge},  
    xlabel style={font=\huge},  
    xticklabel style={font=\huge},
    yticklabel style={font=\huge},
    yticklabel style={font=\huge},
            yticklabel style={
            /pgf/number format/.cd,
            fixed,
            fixed zerofill,
            precision=0  
    },
    legend style={font=\huge, draw=none, fill opacity=0.7, text opacity=1},
    legend pos=south east,
    xmin=-50, xmax=1050,
    ymin=30, ymax=110,
    xtick={100,300,500,700,900},
    ytick={40, 60, 80, 100},
    xtick pos=bottom,    
    ytick pos=left,      
    grid=major,           
    axis background/.style={
        fill={rgb, 255:red,228;green,228;blue,237},
        fill opacity=0.7  
    },
    grid style={color=white},  
    axis line style={color=white},  
    tick style={color=white},  
    cycle list name=color list,
    every axis plot/.append style={thick}
]

\addplot+[
    color={rgb,255:red,31; green,119; blue,180}, 
    line width=2pt,
] coordinates {
    (1, 48.900002) (10, 72.56667) (50, 69.96667) (100, 76.933336) (150, 77.06667) (200, 70.26667) (250, 76.133335) (300, 86.033334) (350, 73.333337) (400, 93.06667) (450, 95.866668) (500, 96.200001) (550, 96.400005) (600, 99.000001) (650, 100) (700, 100) (750, 100) (800, 100) (850, 100) (900, 100) (950, 100) (1000, 100)
};
\addlegendentry{1.30\%}

\addplot+[
    color={rgb,255:red,44; green,160; blue,44}, 
    line width=2pt,
] coordinates {
    (1, 72.0) (10, 78.7) (50, 81.866668) (100, 80.066671) (150, 83.800002) (200, 85.800004) (250, 87.766671) (300, 91.933336) (350, 91.766669) (400, 92.300004) (450, 90.233338) (500, 93.466669) (550, 95.866668) (600, 95.033338) (650, 98.233334) (700, 100) (750, 100) (800, 100) (850, 100) (900, 100) (950, 100) (1000, 100)
};
\addlegendentry{2.60\%}

\addplot+[
    color={rgb,255:red,214; green,39; blue,40}, 
    line width=2pt,
] coordinates {
    (1, 62.200002) (10, 94.100002) (50, 99.666669) (100, 100) (150, 100) (200, 100) (250, 100) (300, 95.700004) (350, 99.633334) (400, 100) (450, 100) (500, 100) (550, 100) (600, 100) (650, 100) (700, 99.966669) (750, 100) (800, 100) (850, 100) (900, 100) (950, 100) (1000, 100)
};
\addlegendentry{5.20\%}

\end{axis}
\end{tikzpicture}
            }
    \label{subfig:cora-running-epoch-asr}
    \vskip -0.1in
    \caption{Cora}
    \end{subfigure}
    &
    \begin{subfigure}{0.25\linewidth}
    \centering
            \resizebox{1.0\linewidth}{!}{
                \begin{tikzpicture}
\begin{axis}[
    xlabel={Epoch},
    ylabel={ASR (\%)},
    ylabel style={font=\huge},  
    xlabel style={font=\huge},  
    xticklabel style={font=\huge},
    yticklabel style={font=\huge},
            yticklabel style={
            /pgf/number format/.cd,
            fixed,
            fixed zerofill,
            precision=0  
        },
    legend style={font=\huge, draw=none, fill opacity=0.7, text opacity=1},
    legend pos=south east,
    xmin=-50, xmax=1050,
    ymin=30, ymax=110,
    xtick={100,300,500,700,900},
    ytick={40, 60, 80, 100},
    xtick pos=bottom,    
    ytick pos=left,      
    grid=major,           
    axis background/.style={
        fill={rgb, 255:red,228;green,228;blue,237},
        fill opacity=0.7  
    },
    grid style={color=white},  
    axis line style={color=white},  
    tick style={color=white},  
    cycle list name=color list,
    every axis plot/.append style={thick}
]

\addplot+[
    color={rgb,255:red,31; green,119; blue,180}, 
    line width=2pt,
] coordinates {
    (1, 64.893) (10, 73.99) (50, 78.938) (100, 85.79) (150, 94.1257) (200, 100) (250, 100) (300, 100) (350, 100) (400, 100) (450, 100) (500, 100) (550, 100) (600, 100) (650, 100) (700, 100) (750, 100) (800, 100) (850, 100) (900, 100) (950, 100) (1000, 100)
};
\addlegendentry{0.90\%}

\addplot+[
    color={rgb,255:red,44; green,160; blue,44}, 
    line width=2pt,
] coordinates {
    (1, 34.645) (10, 53.645) (50, 73.245) (100, 87.923) (150, 95.234) (200, 100) (250, 100) (300, 100) (350, 100) (400, 100) (450, 100) (500, 100) (550, 100) (600, 100) (650, 100) (700, 100) (750, 100) (800, 100) (850, 100) (900, 100) (950, 100) (1000, 100)
};
\addlegendentry{1.80\%}

\addplot+[
    color={rgb,255:red,214; green,39; blue,40}, 
    line width=2pt,
] coordinates {
    (1, 66.7) (10, 83.456) (50, 99.32) (100, 100) (150, 100) (200, 100) (250, 100) (300, 100) (350, 100) (400, 100) (450, 100) (500, 100) (550, 100) (600, 100) (650, 100) (700, 100) (750, 100) (800, 100) (850, 100) (900, 100) (950, 100) (1000, 100)
};
\addlegendentry{3.60\%}

\end{axis}
\end{tikzpicture}
            }
    \label{subfig:citeseer-running-epoch-asr}
    \vskip -0.1in
    \caption{Citeseer}
    \end{subfigure}
    &
    \begin{subfigure}{0.25\linewidth}
    \centering
            \resizebox{1.0\linewidth}{!}{
                \begin{tikzpicture}
\begin{axis}[
    xlabel={Epoch},
    ylabel={ASR (\%)},
    ylabel style={font=\huge},  
    xlabel style={font=\huge},  
    xticklabel style={font=\huge},
    yticklabel style={font=\huge},
    yticklabel style={font=\huge},
            yticklabel style={
            /pgf/number format/.cd,
            fixed,
            fixed zerofill,
            precision=0  
    },
    legend style={font=\huge, draw=none, fill opacity=0.7, text opacity=1},
    legend pos=south east,
    xtick pos=bottom,    
    ytick pos=left,      
    grid=major,           
    axis background/.style={
        fill={rgb, 255:red,228;green,228;blue,237},
        fill opacity=0.7  
    },
    grid style={color=white},  
    axis line style={color=white},  
    tick style={color=white},  
    xmin=-50, xmax=1050,
    ymin=-5, ymax=115,
    xtick={100,300,500,700,900},
    ytick={10, 40, 70, 100},
    cycle list name=color list,
    every axis plot/.append style={thick}
]

\addplot+[
    color={rgb,255:red,31; green,119; blue,180}, 
    line width=2pt,
] coordinates {
    (1, 33.331839) (10, 67.195506) (50, 90.997012) (100, 95.988049) (150, 99.995518) (200, 99.816251) (250, 99.980579) (300, 99.995518) (350, 99.983567) (400, 100) (450, 100) (500, 99.998506) (550, 99.997012) (600, 99.998506) (650, 100) (700, 100) (750, 99.998506) (800, 99.953689) (850, 99.994024) (900, 99.994024) (950, 100) (1000, 100)
};
\addlegendentry{0.10\%}

\addplot+[
    color={rgb,255:red,44; green,160; blue,44}, 
    line width=2pt,
] coordinates {
    (1, 0) (10, 12.955079) (50, 16.444823) (100, 19.573045) (150, 22.58773) (200, 29.739016) (250, 32.103855) (300, 31.863339) (350, 35.934208) (400, 46.642465) (450, 46.176369) (500, 48.989378) (550, 54.827528) (600, 61.554549) (650, 64.149449) (700, 69.458761) (750, 68.483245) (800, 78.652207) (850, 94.02441) (900, 99.247076) (950, 96.517724) (1000, 100)
};
\addlegendentry{0.50\%}

\addplot+[
    color={rgb,255:red,214; green,39; blue,40}, 
    line width=2pt,
] coordinates {
    (1, 13.134346) (10, 44.55101) (50, 52.783878) (100, 70.014491) (150, 73.00826) (200, 73.891155) (250, 77.085108) (300, 77.639343) (350, 79.965341) (400, 80.237231) (450, 81.284452) (500, 81.383051) (550, 84.355907) (600, 87.930804) (650, 89.239458) (700, 89.276805) (750, 93.532918) (800, 95.897758) (850, 97.057023) (900, 99.111132) (950, 99.526433) (1000, 100)
};
\addlegendentry{1.00\% }

\end{axis}
\end{tikzpicture}
            }
    \label{subfig:flickr-running-epoch-asr}
    \vskip -0.1in
    \caption{Flickr}
    \end{subfigure}
    &
    \begin{subfigure}{0.25\linewidth}
    \centering
            \resizebox{1.0\linewidth}{!}{
                \begin{tikzpicture}
\begin{axis}[
    xlabel={Epoch},
    ylabel={ASR (\%)},
    ylabel style={font=\huge},  
    xlabel style={font=\huge},  
    xticklabel style={font=\huge},
    yticklabel style={font=\huge},
    yticklabel style={font=\huge},
            yticklabel style={
            /pgf/number format/.cd,
            fixed,
            fixed zerofill,
            precision=0  
        },
    legend style={font=\huge, draw=none, fill opacity=0.7, text opacity=1},
    legend pos=south east,
    xtick pos=bottom,    
    ytick pos=left,      
    grid=major,           
    axis background/.style={
        fill={rgb, 255:red,228;green,228;blue,237},
        fill opacity=0.7  
    },
    grid style={color=white},  
    axis line style={color=white},  
    tick style={color=white},  
    xmin=-50, xmax=1050,
    xtick={100,300,500,700,900},
    ymin=30, ymax=110,
    ytick={40, 60, 80, 100},
    cycle list name=color list,
    every axis plot/.append style={thick}
]

\addplot+[
    color={rgb,255:red,31; green,119; blue,180}, 
    line width=2pt,
] coordinates {
    (1, 19.638555) (10, 36.462087) (50, 48.259656) (100, 89.856023) (150, 92.880119) (200, 99.921685) (250, 99.918069) (300, 99.913251) (350, 99.922887) (400, 99.871685) (450, 99.838553) (500, 99.932528) (550, 99.741566) (600, 99.801807) (650, 99.835541) (700, 99.831325) (750, 99.901805) (800, 99.910237) (850, 99.888553) (900, 99.918069) (950, 99.937346) (1000, 99.999994)
};
\addlegendentry{0.05\%}

\addplot+[
    color={rgb,255:red,44; green,160; blue,44}, 
    line width=2pt,
] coordinates {
    (1, 19.575905) (10, 59.608435) (50, 79.972888) (100, 89.936142) (150, 99.925297) (200, 99.854819) (250, 99.956022) (300, 99.931926) (350, 99.886143) (400, 99.984334) (450, 99.919275) (500, 99.921081) (550, 99.024105) (600, 99.565063) (650, 99.931924) (700, 99.984334) (750, 99.981924) (800, 99.984936) (850, 99.869275) (900, 99.919275) (950, 99.999994) (1000, 99.999994)
};
\addlegendentry{0.10\%}

\addplot+[
    color={rgb,255:red,214; green,39; blue,40}, 
    line width=2pt,
] coordinates {
    (1, 13.962719) (10, 27.881349) (50, 49.892167) (100, 69.766264) (150, 92.396992) (200, 98.803023) (250, 99.326511) (300, 99.592169) (350, 99.902407) (400, 99.695782) (450, 99.445184) (500, 99.252415) (550, 99.027117) (600, 99.482532) (650, 98.890974) (700, 99.798795) (750, 99.527715) (800, 99.889757) (850, 97.169308) (900, 99.583133) (950, 98.207849) (1000, 99.999994)
};
\addlegendentry{0.20\%}

\end{axis}
\end{tikzpicture}
            }
    \label{subfig:reddit-running-epoch-asr}
    \vskip -0.1in
    \caption{Reddit}
    \end{subfigure}
    \end{tabular}
    \vskip -0.1in
    \caption{{Study on Condensation Epochs.}}
    \label{fig:main-run-epochs}
    \vskip -0.2in
\end{figure*}

\subsection{Ablation Study on Poisoned Node Selection}
In this section, we investigate the effects of the poisoned node selection module. To demonstrate the effectiveness of the selection module, we devise a variant of BGC by replacing the selection module by randomly selecting nodes to attach triggers and assign target labels. We denote the variant by $\text{BGC}_{\text{Rand}}$. The results with the condensation method DC-Graph on the dataset Flickr are presented in Figure~\ref{fig:main-random-select-ablation}. We can observe that $\text{BGC}_{\text{Rand}}$ is inferior to BGC in various settings across different datasets, regarding both {CTA} and {ASR}, demonstrating that \textbf{the selection module can effectively enlarge the attack performance and the utility performance of backdoored GNNs within a limited number of poisoned nodes.} Besides, the standard deviations of BGC are also smaller than $\text{BGC}_{\text{Rand}}$, indicating that selecting the representative nodes can also stabilize the attack performance.  

\begin{table}[]
\caption{{Ablation Study on Trigger Generator}.}
\centering
\resizebox{0.8\linewidth}{!}{ 
\begin{tabular}{|c|c|c|cc|}
\hline
{Generator} & {Dataset} & {Ratio (r)} & {CTA} & {ASR} \\ \hline
\multirow{6}{*}{{MLP}} & \multirow{3}{*}{{Cora}} & {1.30\%} & {81.23} & {100} \\
& & {2.60\%} & {80.67} & {100} \\
& & {5.20\%} & {80.70} & {100} \\ \cline{2-5} 
& \multirow{3}{*}{{Citeseer}} & {0.90\%} & {71.57} & {100} \\
& & {1.80\%} & {71.03} & {100} \\
& & {3.60\%} & {70.60} & {100} \\ \hline
\multirow{6}{*}{{GCN}} & \multirow{3}{*}{{Cora}} & {1.30\%} & {81.55} & {100} \\
& & {2.60\%} & {80.98} & {100} \\
& & {5.20\%} & {81.33} & {100} \\ \cline{2-5} 
& \multirow{3}{*}{{Citeseer}} & {0.90\%} & {72.29} & {100} \\
& & {1.80\%} & {71.57} & {100} \\
& & {3.60\%} & {\textbf{71.32}} & {100} \\ \hline
\multirow{6}{*}{{Transformer}} & \multirow{3}{*}{{Cora}} & {1.30\%} & {\textbf{81.71}} & {100} \\
& & {2.60\%} & {\textbf{81.10}} & {100} \\
& & {5.20\%} & {81.07} & {100} \\ \cline{2-5} 
& \multirow{3}{*}{{Citeseer}} & {0.90\%} & {\textbf{72.34}} & {100} \\
& & {1.80\%} & {\textbf{71.62}} & {100} \\
& & {3.60\%} & {\textbf{71.58}} & {100} \\ \hline
\end{tabular}
}
\label{tab:ablastion-study-trigger-generator}
\vskip -0.2in
\end{table}

\subsection{{Ablation Study on Trigger Generator}}
{To analyze the effect of trigger generator on the attack performance, we replace MLP with a GCN (2 layers) and 1-layer Transformer (1 layer, 8 heads), respectively, for trigger generation. We conduct experiments on Cora and Citeseer with GCond as the condensation method. As shown in Table~\ref{tab:ablastion-study-trigger-generator}, transformer consistently achieves the highest CTA in most configurations (e.g., Cora: 81.71\% at r=1.30\%, outperforming GCN (81.55\%) and MLP (81.23\%) in most cases. Nevertheless, they perform the same regarding CTA scores and the divergence of ASR between generators is marginal.} 

\subsection{{Ablation Study on Directed Attack}}
{To assess the adaptability of BGC to directed attacks, we developed a variant that exclusively poisons a specific class and targets this class during testing. The objective is to induce the backdoored GNN to misclassify nodes from the poisoned class into a designated target class. The results, presented in Table~\ref{tab:ablastion-study-directed-attack}, indicate that this variant achieves attack success rates (ASR) comparable to the original BGC, albeit with slightly reduced clean test accuracy (CTA). This performance discrepancy is likely due to the extensive backdooring of the poisoned class, which disrupts the class distribution.}

\begin{table}[]
\caption{{Ablation Study on Directed Attack}.}
\centering
\resizebox{0.85\linewidth}{!}{ 
\begin{tabular}{|c|c|c|cc|}
\hline
{Method} & {Dataset} & {Ratio (r)} & {CTA} & {ASR} \\ \hline
\multirow{6}{*}{{BGC}} & \multirow{3}{*}{{Cora}} & {1.30\%} & {\textbf{81.23}} & {100} \\
& & {2.60\%} & {\textbf{80.67}} & {100} \\
& & {5.20\%} & {\textbf{80.70}} & {100} \\ \cline{2-5} 
& \multirow{3}{*}{{Citeseer}} & {0.90\%} & {\textbf{71.57}} & {100} \\
& & {1.80\%} & {\textbf{71.03}} & {100} \\
& & {3.60\%} & {\textbf{70.60}} & {100} \\ \hline
\multirow{6}{*}{{Directed}} & \multirow{3}{*}{{Cora}} & {1.30\%} & {81.03} & {100} \\
& & {2.60\%} & {80.64} & {100} \\
& & {5.20\%} & {80.69} & {100} \\ \cline{2-5} 
& \multirow{3}{*}{{Citeseer}} & {0.90\%} & {71.55} & {100} \\
& & {1.80\%} & {70.87} & {100} \\
& & {3.60\%} & {70.55} & {100} \\ \hline
\end{tabular}
}
\label{tab:ablastion-study-directed-attack}
\vskip -0.2in
\end{table}

\subsection{Hyper-parameter Analysis}
\textbf{Number of GNNs' Layers.} Here, we aim to understand how the number of GNN layers affects the attack performance and model utility. We carry out experiments on three datasets: Cora, Citeseer, and Flickr, utilizing the condensation method GCond. The results are presented in Table~\ref{tab:layer-num}, from which we can observe that the layer number does not necessarily influence the {CTA} scores a lot, except on dataset Cora. On dataset Cora, the GNN's utility outperforms the other two settings when the layer number is set to $2$. On the other hand, the change of layer number also does not affect the {ASR} scores and on the dataset Flickr, the {ASR} decreases slightly.

\textbf{Condensation Epochs.} We further investigate the impact of condensation epochs on attack and utility performance. Since the number of condensation epochs has a great influence on the quality of the condensed graph, we report the attack and utility performance by varying condensation epochs from $50$ to $1000$ on four datasets, using the condensation method GCond. As depicted in Figure~\ref{fig:main-run-epochs}, ASR scores first increases and converges to a stable range of values.
\begin{table}
\centering
\caption{Study on Poisoning Ratio. $r$ denotes the condensation ratio, \textit{P. R.} denotes the poison ratio and \textit{P. N.} denotes the poison number.}
\vskip -0.05in
\label{tab:poisoning-ratio}
\scalebox{0.97}{
\begin{tabular}{|c|c|cc|cc|cc|}
\hline
\multirow{2}{*}{Datasets} &
  \multirow{2}{*}{P. R.} &
  \multicolumn{2}{c|}{DC-Graph} &
  \multicolumn{2}{c|}{GCond} &
  \multicolumn{2}{c|}{GCond-X} \\ \cline{3-8} 
                           &                       & CTA     & ASR     & CTA     & ASR     & CTA     & ASR     \\ \hline
\multirow{3}{*}{\begin{tabular}{@{}c@{}} Cora, \\ $r=1.30\%$ \end{tabular}}& \multirow{1}{*}{0.10}  & \textbf{75.90} & \textbf{100.0}   & \textbf{81.23} & \textbf{100.0}   & \textbf{76.30} & \textbf{100.0}   \\
                           & \multirow{1}{*}{0.15} & 75.26& 100.0  & 80.06 & 100.0   & 75.30 & 100.0   \\
                           & \multirow{1}{*}{0.20}  & 75.13 & 99.93 & 78.33  & 100.0  & 75.07 & 100.0   \\ \hline
\multirow{3}{*}{\begin{tabular}{@{}c@{}} Citeseer, \\ $r=3.60\%$ \end{tabular}}& \multirow{1}{*}{0.10}  & 68.83 & 100.0   & 69.03 & 100.0   & 67.27 & 100.0   \\
                           & \multirow{1}{*}{0.15} & \textbf{70.00}& \textbf{100.0}  & \textbf{70.60} & \textbf{100.0}   & \textbf{72.13} & \textbf{100.0}   \\
                           & \multirow{1}{*}{0.20}  & 66.40 & 100.0 & 69.27  & 100.0  & 70.83 & 99.40   \\ \hline\hline
\multirow{2}{*}{Datasets} &
  \multirow{2}{*}{P. N.} &
  \multicolumn{2}{c|}{DC-Graph} &
  \multicolumn{2}{c|}{GCond} &
  \multicolumn{2}{c|}{GCond-X} \\ \cline{3-8}
                           &                       & CTA     & ASR     & CTA     & ASR     & CTA     & ASR     \\ \hline
\multirow{3}{*}{\begin{tabular}{@{}c@{}} Flickr, \\ $r=0.10\%$ \end{tabular}}& \multirow{1}{*}{60}   & 45.86  & 99.56 & 45.38 & 99.13 & 44.85 & \textbf{100.0} \\
                           & \multirow{1}{*}{80}   & \textbf{46.48} & 99.98 & \textbf{46.54} & \textbf{99.83} & 46.15 & 98.26 \\
                           & \multirow{1}{*}{100}   & 46.18 & \textbf{100.0} & 46.18 & 99.66 & \textbf{46.19} & 97.21 \\ \hline
\multirow{3}{*}{\begin{tabular}{@{}c@{}} Reddit, \\ $r=0.05\%$ \end{tabular}}& \multirow{1}{*}{130}   & \textbf{85.74}  & 96.12 & \textbf{88.50} & 97.14 & \textbf{88.10} & 96.15 \\
                           & \multirow{1}{*}{180}   & 85.40 & \textbf{99.90} & 88.37 & \textbf{99.84} & 88.04 & \textbf{99.89} \\
                           & \multirow{1}{*}{230}   & 85.38 & 98.29 & 88.14 & 98.92 & 87.47 & 99.10 \\ \hline
\end{tabular}
}
\vskip -0.22in
\end{table}
\begin{table*}[]
\centering
\vskip -0.03in
\caption{Study on the Number of GNNs' Layers. $l$ denotes the layer number.}
\vskip -0.07in
\resizebox{1.0\linewidth}{!}{ 
\begin{tabular}{|cc|ccc|ccc|ccc|}
\hline
\multicolumn{2}{|c|}{Datasets} &
  \multicolumn{3}{c|}{Cora} &
  \multicolumn{3}{c|}{Citeseer} &
  \multicolumn{3}{c|}{Flickr} \\ \hline
\multicolumn{2}{|c|}{Ratio (r)} &
  1.30\% &
  2.60\% &
  5.20\% &
  0.90\% &
  1.80\% &
  3.60\% &
  0.10\% &
  0.50\% &
  1.00\% \\ \hline
\multirow{2}{*}{$l$ = 1} &
  CTA &
  66.53 (2.24) &
  72.93 (1.75) &
  76.47 (0.52) &
  69.53 (0.33) &
  \textbf{69.43 (0.96)} &
  68.23 (0.12) &
  46.43 (0.21) &
  \textbf{47.00 (0.04)} &
  \textbf{47.12 (0.17)} \\
 &
  ASR &
  100.0 (0.00) &
  100.0 (0.00) &
  100.0 (0.00) &
  {100.0 (0.00)} &
  \textbf{100.0 (0.00)} &
  100.0 (0.00) &
  {100.0 (0.00)} &
  \textbf{100.0 (0.00)} &
  \textbf{100.0 (0.00)} \\ \hline
\multirow{2}{*}{$l$ = 2} &
  CTA &
  \textbf{75.90 (0.94)} &
  \textbf{75.00 (0.51)} &
  \textbf{78.43 (0.31)} &
  \textbf{70.27 (0.50)} &
  69.37 (0.94) &
  \textbf{70.00 (0.22)} &
  \textbf{46.48 (0.21)} &
  46.44 (0.13) &
  46.77 (0.30) \\
 &
  ASR &
  \textbf{100.0 (0.00)} &
  \textbf{100.0 (0.00)} &
  \textbf{100.0 (0.00)} &
  \textbf{100.0 (0.00)} &
  100.0 (0.00) &
  \textbf{100.0 (0.00)} &
  \textbf{100.0 (0.00)} &
  99.25 (0.65) &
  99.11 (0.40) \\ \hline
\multirow{2}{*}{$l$ = 3} &
  CTA &
  69.07 (1.70) &
  73.87 (0.84) &
  76.77 (0.47) &
  69.63 (0.52) &
  68.60 (0.65) &
  68.23 (0.21) &
  46.28 (0.31) &
  46.17 (0.22) &
  46.89 (1.27) \\
 &
  ASR &
  100.0 (0.00) &
  100.0 (0.00) &
  100.0 (0.00) &
  100.0 (0.00) &
  100.0 (0.00) &
  100.0 (0.00) &
  99.01 (0.21) &
  99.91 (0.03) &
  99.28 (0.34) \\ \hline
\end{tabular}
}
\label{tab:layer-num}
\vskip -0.2in
\end{table*}

\textbf{Varying the Poisoning Ratio}
We explore the effect of the poisoning ratio on the attack and utility performance.
We report the results on four datasets in Table~\ref{tab:poisoning-ratio}. As shown in the table, in all cases for different datasets, a larger poisoning ratio does not necessarily lead to better utility performance. In contrast, larger poisoning ratio can lead to worse GNN utility. For instance, given the dataset Reddit, the {CTA} scores across 3 condensation methods monotonically decease as the poison number increases. This suggests that a higher number of poisoned nodes can degrade the quality of the condensed graph, thereby reducing the utility of the backdoored GNN. 

\textbf{Various Trigger Sizes}
As shown in previous work~\cite{ndss2023doorping,dai2023www-unnoticeable}, larger trigger sizes can contribute to higher attack performance.
To investigate the effect of trigger size on the attack and utility performance, we experiment on the dataset Flickr with two condensation methods and report the results under three different condensation ratios in Figure~\ref{fig:main-trigger}. As depicted in the figure, the {ASR} scores are all close to $100\%$ and increase as the trigger size becomes larger. In contrast, the {CTA} scores decline as the trigger size increases. Despite the negative effects on {CTA} scores caused by increasing the trigger size, the degradation is marginal and the utility performance is still acceptable. For example, the declines of {CTA} scores in three settings are only 0.51\%, 0.94\%, and 0.92\%, respectively. Therefore, increasing the size of triggers can cause a trade-off between attack performance and GNN utility.

\section{Related Work}
\label{sec:related-work}
\textbf{Graph Condensation.} Recently, graph condensation~\cite{gcond2024surveyEmory,gcond2024surveyUQ,gcond2024surveyZJU,opengraph24xlh,www24LLMRec} is a technique for efficient graph learning, which aims to synthesize small graph datasets from large ones. GCond~\cite{iclr2022gcond} is the first graph condensation method and focuses on node-level datasets. It matches the training gradient of the original datasets with condensed datasets to achieve comparable performance. Doscond~\cite{doscond-kdd2022} extends the gradient matching paradigm to the graph-level datasets and proposes a one-step update strategy to enhance the condensation efficiency. GCSR~\cite{kdd2024GCSR} and SGDD~\cite{nips2023does} are proposed to incorporate more comprehensive graph structure information into the condensed datasets.  CTRL~\cite{arXiv2024NTU-CTRL} offers a better initialization and a more refined strategy for gradient matching. GEOM~\cite{arXiv2024ntu-geom} proposes to utilize a well-trained expert trajectory to supervise the condensation. FGD~\cite{nips2023FGD} aims at improving group fairness for node classification task in graph condensation.  MCond\cite{icde2024MCond} enables efficient inductive inference. On the other hand, CaT~\cite{icdm2023CaT} and GCDM~\cite{arXiv2022GCDM} both adopt maximum mean discrepancy to match the distribution between original graphs and condensed graphs. EXGC~\cite{fang2024exgcWebConference} accelerates condensation via Mean-field variational approximation and inject explainability via existing explanation techniques. Different from the aforementioned methods that formulate condensation as a bi-level optimization problem, MIRAGE~\cite{iclr2024mirage} investigates graph condensation from a heuristic perspective and proposes a model-agnostic method. Besides, GC-SNTK~\cite{wangling2023fastGC}, LiteGNTK~\cite{kdd2023LiteGNTK} and SFGC~\cite{sfgc2023nips} all adopt the Kernel Ridge Regression formalization while LiteGNTK focuses on graph-level tasks and the other two are devised for node-level tasks. 
Several surveys have been released recently and more comprehensive introduction can be found in ~\cite{gcond2024surveyUQ,gcond2024surveyZJU,gcond2024surveyEmory}.
While those works focus on pursuing an optimal trade-off between GNN utility and condensation ratio, the security risks  stemming from the condensation process are overlooked and haven't been investigated. Thus, we first explore this issue and propose an effective backdoor attack method against graph condensation.

\textbf{Backdoor Attacks on Graph.} Graph backdoor attacks targets graph neural networks (GNNs) by embedding specific triggers during training, causing the GNN to make malicious predictions when these triggers are detected, while performing normally on clean inputs~\cite{icml2019evasionGraphAttack,poisonattack2018kdd,wenqi23TKDE-CopyAttack}. According to the categorization by the stages the attack occurs~\cite{dai2023www-unnoticeable}, adversarial attacks on GNNs mainly contain three types: poisoning attack~\cite{poisonattack2018kdd,poisonAttack2020www}, evasion attack~\cite{aaai2020evasionGraphAttack,icml2019evasionGraphAttack}, and backdoor attack~\cite{dai2023www-unnoticeable,gta2021usenix}. In this paper, we focus on the backdoor attacks. The graph backdoor attack~\cite{raid2022TransGBA,CCS2022Poster,WiseML2021Explain,ijcai2023WorkShopPerCBA,corr2023SemanticBGA,zjiahao24wwwLinearRec} is a training time attack~\cite{ndss2023doorping}. It injects a hidden backdoor into the target GNNs via a backdoored training graph. At the test time, the successfully backdoored GNNs perform well on the clean test samples but misbehave on the triggered samples. The first work~\cite{sacmat2021BaGNN} on graph backdoor randomly generates graphs as triggers while GBAST~\cite{ICCC2021GBAST} generates triggers based on subgraphs. Different from the universal triggers, GTA~\cite{gta2021usenix} proposes a generator to adaptively obtain sample-specific triggers. To achieve unnoticeable graph backdoor attacks, UGBA~\cite{dai2023www-unnoticeable} limits the attack budget and improves the similarity between triggers and target nodes. GCBA~\cite{icml2023GCBA} studies a new setting where supervisory labels are unavailable and proposes the first backdoor attack against graph contrastive learning. Those efforts focus on injecting triggers into the original graph during model training, which is infeasible to poison the condensed graph data. Recently, two related studies~\cite{ndss2023doorping,iclr2024rethinking} on backdoor attacks against dataset distillation for image data are proposed and reveal the vulnerability of dataset condensation in image domain. However, their design is limited to producing universal triggers for images, which is infeasible for graph.
\section{{Discussion}}
\label{sec:discussions}
In this section, we discuss the difference between backdoor attacks against graph condensation (BGC) and conventional graph backdoor attacks (CGB), challenges to defense attacks against graph condensation and potential defense mechanism, the potential application of BGC in real-world scenario, and the limitations of BGC in more practical settings.

\textbf{Different from Conventional Graph Backdoor.} The goals of BGC and CGB are identical: backdooring the downstream GNN model through injection and manipulating its output while attaching trigger. However, BGC poses significantly greater challenges due to following reasons: 1) In graph condensation, the condensed graph is consistently updated towards the goal of maximizing the utility of downstream GNN, making it intractable to maintain the effectiveness of trigger, whereas in CGB, the trigger is generated on the static original graph. 2) The quality of the condensed graph highly depends on the original graph and the condensation algorithm, complicating the design of injections that do not degrade the condensed graph's quality.
\begin{figure}[]
\centering
{\includegraphics[width=0.9\linewidth]{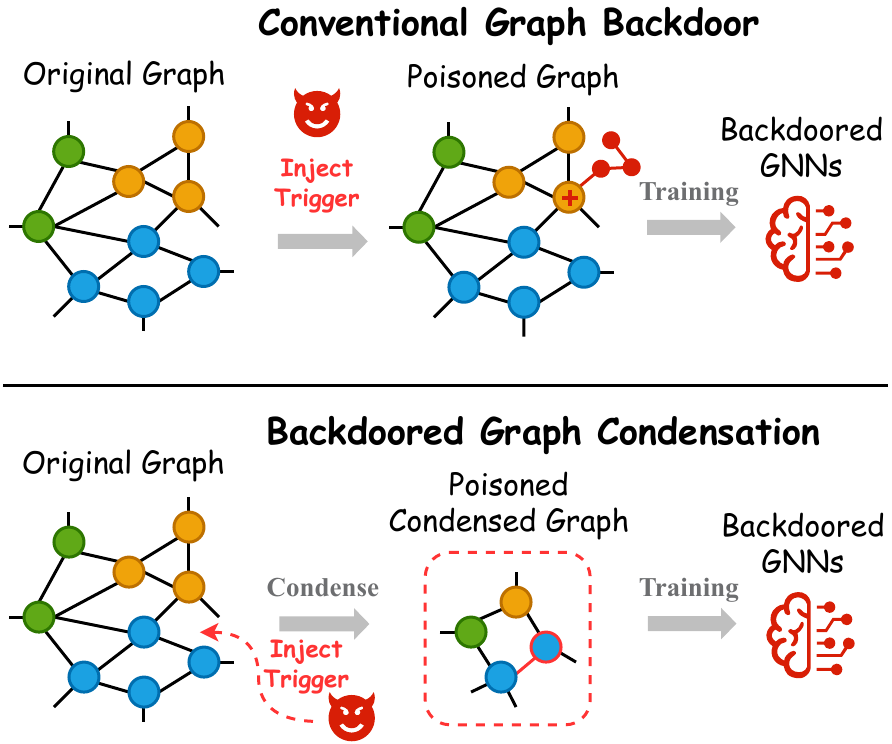}}
\caption{Comparison between Backdoor Graph Condensation (BGC) and Conventional Graph Backdoor (CGB). }
\vskip -0.25in
\label{fig:BGCvsCGC}
\end{figure}

\textbf{More Challenging to Defend BGC.} As shown in Figure~\ref{fig:BGCvsCGC}, different from CGB where the trigger is explicitly attach in the graph for GNN training, the malicious information of triggers is embedded within the synthetic nodes in BGC. Therefore, there is no explicit trigger present in the condensed graph, rendering detection-based and prune-based defense methods ineffective in mitigating attacks against graph condensation. The potentially effective defense mechanism should focus on designing robust GNN training algorithms or designing new GNN's architecture that can mitigate these attacks.

\textbf{Example of Real-world Scenario.} In this work, we conceptualize the attacker as a graph condensation service provider, which can be represented by real-world data companies. In the era of AI, data is indispensable for developing advanced models, and graphs are a key data modality, driving the demand for high-quality graph data and the rise of data provider companies such as Scale AI. While such services offer convenience, it is essential to highlight the potential security vulnerabilities in these scenarios where the data provider may not be trustworthy, or where third parties could inadvertently introduce backdoors. In such cases, if a malicious actor gains control over the model supporting a product service, the resulting economic losses could be catastrophic. Therefore, we attempt to reveal vulnerabilities in graph condensation and the associated risks in this work.
\begin{figure}
    \centering
    \setlength\tabcolsep{1.5pt}
    \begin{tabular}{cccc}
    \begin{subfigure}{0.5\linewidth}
    \centering
        \resizebox{1.0\linewidth}{!}{
            \begin{tikzpicture}
\begin{axis}[
    ylabel={CTA (\%)},
    xmin=0.8, xmax=4.2,  
    ymin=45.75, ymax=47.75,  
    xtick={1,2,3,4,5,6},  
    ytick={46.0, 46.5, 47.0, 47.5},  
    ylabel style={font=\huge},          
    xticklabel style={font=\huge},
    yticklabel style={font=\huge},
    yticklabel style={font=\huge},
            yticklabel style={
            /pgf/number format/.cd,
            fixed,
            fixed zerofill,
            precision=1  
    },
    legend style={font=\huge, draw=none, fill opacity=0.7, text opacity=1},  
    xtick pos=bottom,    
    ytick pos=left,      
    grid=major,  
    axis background/.style={
        fill={rgb,255:red,228;green,228;blue,237},
        fill opacity=0.7  
    },
    grid style={color=white},  
    axis line style={color=white},  
    tick style={color=white},  
    every axis plot/.append style={thick}
]

\addplot[dash pattern=on 10pt off 4pt,color={rgb,255:red,59;green,91;blue,146}, mark=*, mark options={solid, fill opacity=0.75, draw opacity=0.0}, mark size=4.5pt, line width=2pt] 
coordinates {
    (1, 46.479703)
    (2, 46.465631)
    (3, 46.301309)
    (4, 46.19)
};
\addlegendentry{0.10\%}

\addplot[dash pattern=on 10pt off 4pt,color={rgb,255:red,56;green,173;blue,72}, mark=*, mark options={solid, fill opacity=0.75, draw opacity=0.0}, mark size=4.5pt, line width=2pt]
coordinates {
    (1, 47.007)
    (2, 47.026397)
    (3, 46.943777)
    (4, 46.74)
};
\addlegendentry{0.50\%}

\addplot[dash pattern=on 10pt off 4pt,color={rgb,255:red,217;green,84;blue,77}, mark=*, mark options={solid, fill opacity=0.75, draw opacity=0.0}, mark size=4.5pt, line width=2pt]
coordinates {
    (1, 47.186031)
    (2, 47.116031)
    (3, 46.836672)
    (4, 46.65)
};
\addlegendentry{1.00\%}

\end{axis}
\end{tikzpicture}
        }
    \label{subfig:dc-graph-cta}
    \vskip -0.15in
    \end{subfigure}
    &
    \begin{subfigure}{0.5\linewidth}
    \centering
            \resizebox{1.0\linewidth}{!}{
                \begin{tikzpicture}
\begin{axis}[
    ylabel={CTA (\%)},
    xmin=0.8, xmax=4.2,  
    ymin=45.8, ymax=47.4,  
    xtick={1,2,3,4,5,6},  
    ytick={46.0, 46.4, 46.8, 47.2},  
    ylabel style={font=\huge},          
    xticklabel style={font=\huge},
    yticklabel style={font=\huge},
    yticklabel style={font=\huge},
            yticklabel style={
            /pgf/number format/.cd,
            fixed,
            fixed zerofill,
            precision=1  
    },
    legend style={font=\huge, draw=none, fill opacity=0.7, text opacity=1},  
    xtick pos=bottom,    
    ytick pos=left,      
    grid=major,  
    axis background/.style={
        fill={rgb,255:red,228;green,228;blue,237},  
        fill opacity=0.7  
    },
    grid style={color=white},  
    axis line style={color=white},  
    tick style={color=white},  
    every axis plot/.append style={thick}
]

\addplot[dash pattern=on 10pt off 4pt,color={rgb,255:red,59;green,91;blue,146}, mark=*, mark options={solid, fill opacity=0.75, draw opacity=0.0}, mark size=4.5pt, line width=2pt] 
 coordinates {
    (1, 46.606156)
    (2, 46.564145)
    (3, 46.412588)
    (4, 46.146856)
};
\addlegendentry{0.10\%}


\addplot[dash pattern=on 10pt off 4pt,color={rgb,255:red,56;green,173;blue,72}, mark=*, mark options={solid, fill opacity=0.75, draw opacity=0.0}, mark size=4.5pt, line width=2pt]
coordinates {
    (1, 46.823136)
    (2, 46.760483)
    (3, 46.679995)
    (4, 46.54872)
};
\addlegendentry{0.50\%}

\addplot[dash pattern=on 10pt off 4pt,color={rgb,255:red,217;green,84;blue,77}, mark=*, mark options={solid, fill opacity=0.75, draw opacity=0.0}, mark size=4.5pt, line width=2pt]
coordinates {
    (1, 47.004051)
    (2, 46.909964)
    (3, 46.836672)
    (4, 46.702648)
};
\addlegendentry{1.00\%}

\end{axis}
\end{tikzpicture}
            }
    \label{subfig:gcond-cta}
    \vskip -0.15in
    \end{subfigure}
    \\
    \begin{subfigure}{0.5\linewidth}
    \centering
            \resizebox{1.0\linewidth}{!}{
                \begin{tikzpicture}
\begin{axis}[
    xlabel={Trigger Size},
    xlabel style={font=\huge},          
    ylabel={ASR (\%)},
    xmin=0.8, xmax=4.2,  
    ymin=90, ymax=106,  
    xtick={1,2,3,4,5,6},  
    ytick={92, 96, 100, 104},  
    ylabel style={font=\huge},          
    xticklabel style={font=\huge},
    yticklabel style={font=\huge},
    legend style={font=\huge, draw=none, fill opacity=0.7, text opacity=1},  
    xtick pos=bottom,    
    ytick pos=left,      
    grid=major,  
    axis background/.style={
        fill={rgb,255:red,228;green,228;blue,237},
        fill opacity=0.7  
    },
    grid style={color=white},  
    axis line style={color=white},  
    tick style={color=white},  
    every axis plot/.append style={thick}
]

\addplot[dash pattern=on 10pt off 4pt,color={rgb,255:red,59;green,91;blue,146}, mark=*, mark options={solid, fill opacity=0.75, draw opacity=0.0}, mark size=4.5pt, line width=2pt] 
coordinates {
    (1, 96.174129)
    (2, 97.291264)
    (3, 96.425985)
    (4, 97.58)
};
\addlegendentry{0.10\%}

\addplot[dash pattern=on 10pt off 4pt,color={rgb,255:red,56;green,173;blue,72}, mark=*, mark options={solid, fill opacity=0.75, draw opacity=0.0}, mark size=4.5pt, line width=2pt]
coordinates {
    (1, 94.83567)
    (2, 95.862263)
    (3, 99.011222)
    (4, 99.45)
};
\addlegendentry{0.50\%}

\addplot[dash pattern=on 10pt off 4pt,color={rgb,255:red,217;green,84;blue,77}, mark=*, mark options={solid, fill opacity=0.75, draw opacity=0.0}, mark size=4.5pt, line width=2pt]
coordinates {
    (1, 96.427655)
    (2, 97.840721)
    (3, 97.4648412)
    (4, 98.51)
};
\addlegendentry{1.00\%}

\end{axis}
\end{tikzpicture}
            }
    \label{subfig:dc-graph-asr}
    \vskip -0.15in
    \caption{DC-Graph}
    \end{subfigure}
    &
    \begin{subfigure}{0.5\linewidth}
    \centering
            \resizebox{1.0\linewidth}{!}{
                \begin{tikzpicture}
\begin{axis}[
    xlabel={Trigger Size},
    xlabel style={font=\huge},          
    ylabel={ASR (\%)},
    xmin=0.8, xmax=4.2,  
    ymin=97.5, ymax=101.5,  
    xtick={1,2,3,4,5,6},  
    ytick={98, 99, 100, 101},  
    ylabel style={font=\huge},          
    xticklabel style={font=\huge},
    yticklabel style={font=\huge},
    legend style={font=\huge, draw=none, fill opacity=0.7, text opacity=1},  
    xtick pos=bottom,    
    ytick pos=left,      
    grid=major,  
    axis background/.style={
        fill={rgb,255:red,228;green,228;blue,237},  
        fill opacity=0.7  
    },
    grid style={color=white},  
    axis line style={color=white},  
    tick style={color=white},  
    every axis plot/.append style={thick}
]

\addplot[dash pattern=on 10pt off 4pt,color={rgb,255:red,59;green,91;blue,146}, mark=*, mark options={solid, fill opacity=0.75, draw opacity=0.0}, mark size=4.5pt, line width=2pt] 
coordinates {
    (1, 99.380579)
    (2, 99.318146)
    (3, 99.457717)
    (4, 99.825215)
};
\addlegendentry{0.10\%}

\addplot[dash pattern=on 10pt off 4pt,color={rgb,255:red,56;green,173;blue,72}, mark=*, mark options={solid, fill opacity=0.75, draw opacity=0.0}, mark size=4.5pt, line width=2pt]
coordinates {
    (1, 99.076774)
    (2, 99.75061)
    (3, 99.67732)
    (4, 99.96564)
};
\addlegendentry{0.50\%}

\addplot[dash pattern=on 10pt off 4pt,color={rgb,255:red,217;green,84;blue,77}, mark=*, mark options={solid, fill opacity=0.75, draw opacity=0.0}, mark size=4.5pt, line width=2pt]
coordinates {
    (1, 97.904571)
    (2, 98.791035)
    (3, 98.530914)
    (4, 99.472926)
};
\addlegendentry{1.00\%}

\end{axis}
\end{tikzpicture}
            }
    \label{subfig:gcond-asr}
    \vskip -0.15in
    \caption{GCond}
    \end{subfigure}
    \end{tabular}
    \vskip -0.1in
    \caption{Study on Different Trigger Sizes.}
    \label{fig:main-trigger}
    \vskip -0.25in
\end{figure}

{\textbf{Limitations.} In real-world settings~\cite{iclr2024PracticalBackdoor,iclr2025PracticalBackdoor_clean-label}, attacker capabilities are often constrained. We expand on key limitations:}

\begin{itemize}
    \item {\textbf{Partial Data Access}: An attacker might only control a subset of nodes/edges (e.g., in federated or crowdsourced graphs). Our method’s reliance on representative node selection could still apply if the attacker poisons high-influence nodes within their accessible subset. Via the selection and consistent trigger update, our method may still work well and launch effective attack. However, trigger generated with limited data can limit its effectiveness among different other nodes. Therefore, a more generalizable trigger may be desired.}

    \item {\textbf{Process Access}: If the condensation algorithm is hidden (e.g., a commercial API), dynamic trigger optimization becomes infeasible. This setting is totally different from our assumption and predefined scenario. Under this setting, an invariant trigger is required, which can survive across the consistent optimization.}
\end{itemize}
{These scenarios highlight underexplored challenges in backdoor attacks: 1) the tension between localized attack efficacy and global pattern propagation, and 2) the robustness-adaptability trade-off in black-box condensation environments. While addressing these limitations requires substantial methodological innovation (e.g., meta-trigger learning for partial visibility, surrogate condensation modeling for API-obscured settings), we defer such explorations to future research. This work establishes foundational insights for graph backdoor attacks under constrained threat models.}
\section{Conclusion}
\label{sec:conclusion}
In this paper, we first propose the task of backdoor graph condensation. We envision a realistic scenario where the attacker is a malicious graph condensation provider and possesses the information of the graph data. We set two primary objectives to launch successful attacks: 1) the injection of triggers cannot affect the quality of condensed graphs, maintaining the utility of GNNs trained on them; and 2) the effectiveness of triggers should be preserved throughout the condensation process, achieving a high attack success rate. To pursue these two goals, we propose the first backdoor attack against graph condensation (BGC). Specifically, we inject toxic information into the condensed graph by injecting triggers into the original graph during the condensation process, which is different from all previous graph backdoor attacks that perform attacks during the model training stage. Extensive experiments across multiple datasets, GNN architectures, and dataset condensation methods demonstrate that our proposed method achieves impressive attack performance and utility performance. The resilience of our method against the defense mechanisms are also verified via experiments. We hope this work could set the stage for future research in the field of graph condensation security and raise awareness about the related implications among the research community. 





\section{acknowledgements}
This work is supported by  National Key Research and Development Program of China under Grant 2022YFA1004102, and in part by the Guangdong Major Project of Basic and Applied Basic Research under Grant 2023B0303000010. This work has also been supported by Hong Kong Research Grants Council through Theme-based Research Scheme (project no. T43-513/23-N).
\bibliographystyle{IEEEtran}
\bibliography{IEEEabrv,references}

\end{document}